\documentclass{article}

\PassOptionsToPackage{numbers,sort&compress}{natbib}

    \usepackage[final]{neurips_2021}

\usepackage[utf8]{inputenc} %
\usepackage[T1]{fontenc}    %
\usepackage[breaklinks]{hyperref}       %
\usepackage{url}            %
\usepackage{booktabs}       %
\usepackage{amsfonts}       %
\usepackage{nicefrac}       %
\usepackage{microtype}      %
\usepackage[table,dvipsnames]{xcolor}        %

\usepackage[normalem]{ulem}
\useunder{\uline}{\ul}{}

\usepackage{amsmath}
\usepackage{amssymb}
\usepackage{amsthm}
\usepackage{multicol}
\usepackage{epsfig}
\usepackage{wrapfig}

\usepackage{microtype}
\usepackage{graphicx}
\usepackage{subfigure}
\usepackage{booktabs} %

\hypersetup{breaklinks=true}

\DeclareUnicodeCharacter{2212}{-}

\usepackage{microtype}
\usepackage{graphicx}
\usepackage{booktabs} %
\usepackage{multirow}
\usepackage{siunitx}
\usepackage[frozencache=true,cachedir=.]{minted}

\usepackage{graphicx}

\usepackage{tikz}
\usepackage{pgfplots}
\pgfplotsset{compat=newest}
\DeclareUnicodeCharacter{2212}{−}
\usepgfplotslibrary{groupplots,dateplot}
\usetikzlibrary{patterns,shapes.arrows}
\usetikzlibrary{decorations.text,calc,shapes,arrows,arrows.meta, positioning,shapes.misc,decorations.markings,decorations.markings,decorations.pathreplacing,matrix,spy}
\usepgfplotslibrary{groupplots}
\usepgfplotslibrary{dateplot}

\newcommand{\bind}{\otimes}

\newcommand{\inv}{\dagger}
\newcommand{\invA}{*}

\newcommand{\bigO}{\mathcal{O}}

\usepackage{caption}
\usepackage{multirow}
\usepackage{appendix}

\usepackage[disable,textsize=tiny]{todonotes}

\DeclareMathOperator*{\argmax}{arg\,max}

\usepackage[autostyle, english=american]{csquotes}
\MakeOuterQuote{"}

\usepackage{adjustbox}

\usepackage{cleveref}
\crefformat{section}{\S#2#1#3} %
\crefformat{subsection}{\S#2#1#3}
\crefformat{subsubsection}{\S#2#1#3}
\crefname{figure}{Fig.}{}

\usepackage{breakurl}

\title{Learning with Holographic Reduced Representations }

\author{
Ashwinkumar Ganesan\textsuperscript{1}\thanks{Research work completed prior to joining Amazon.},
Hang Gao\textsuperscript{1}, 
Sunil Gandhi\textsuperscript{1}\footnotemark[1],
Edward Raff\textsuperscript{1,2,3},
Tim Oates\textsuperscript{1}\\
\textbf{James Holt\textsuperscript{2},
Mark McLean\textsuperscript{2}}\\
\textsuperscript{1}University of Maryland, Baltimore County \\
\textsuperscript{2}Laboratory for Physical Sciences \\
\textsuperscript{3}Booz Allen Hamilton
}

\begin{document}

\maketitle

\begin{abstract}
Holographic Reduced Representations (HRR) are a method for performing symbolic AI on top of real-valued vectors \cite{Plate1995} by associating each vector with an abstract concept, and providing mathematical operations to manipulate vectors as if they were classic symbolic objects. This method has seen little use outside of older symbolic AI work and cognitive science. Our goal is to revisit this approach to understand if it is viable for enabling a hybrid neural-symbolic  approach to learning as a differentiable component of a deep learning architecture. HRRs today are not effective in a differentiable solution due to numerical instability, a problem we solve by introducing a projection step that forces the vectors to exist in a well behaved point in space. In doing so we improve the concept retrieval efficacy of HRRs by over $100\times$. Using multi-label classification we demonstrate how to leverage the symbolic HRR properties to develop an output layer and loss function that is able to learn effectively, and allows us to investigate some of the pros and cons of an HRR neuro-symbolic learning approach. 
Our code can be found at \url{https://github.com/NeuromorphicComputationResearchProgram/Learning-with-Holographic-Reduced-Representations}
\end{abstract}

\section{Introduction}
\label{sec:introduction}

Symbolic and connectionist (or "neural") based approaches to Artificial Intelligence (AI) and Machine Learning (ML) have often been treated as two separate, independent methods of approaching AI/ML. This does not need to be the case, and our paper proposes to study the viability of a hybrid approach to propagation based learning. 
In particular, we make use of the Holographic Reduced Representation (HRR) approach originally proposed by \citet{Plate1995}. Plate proposed using circular convolution as a "binding" operator. Given two vectors in a $d$ dimensional feature space,  $\boldsymbol{x}, \boldsymbol{y} \in \mathbb{R}^d$, they can be ``bound'' together using circular convolution, which we denote as $\boldsymbol{s} = \boldsymbol{x} \bind \boldsymbol{y}$. This gives us a new result $\boldsymbol{s} \in \mathbb{R}^d$. The HRR approach also includes an inversion operation $\inv$ that maps $\mathbb{R}^d \to \mathbb{R}^d$. 

With this binding and inverse operation, Plate showed that we can assign vectors to have a conceptual symbolic meaning and construct \textit{statements} (or "sentences") from these representations. For example, we can construct $S = \mathit{red} \bind \mathit{cat} + \mathit{blue} \bind \mathit{dog} $ to represent ``a red cat and a blue dog''. HRRs can also \textit{query} the statement representation $S$. To ask which animal was red, we compose $S \bind \mathit{red}^\inv \approx \mathit{cat}$ which gives us a numeric output approximately equal to the vector representing "cat". 

The HRR framework requires a constant amount of memory, relies on well-optimized and scalable operations like the Fast Fourier Transform (FFT), provides symbolic manipulation, and uses only differentiable operations, that would make it seem ideal as a tool for exploring neuro-symbolic modeling. Unfortunately back-propagating through HRR operations does not learn in practice, rendering it seemingly moot for such neuro-symbolic research. The goal of this work is to find sufficient conditions for successful back-propagation based learning with the HRR framework, and to develop evidence of its potential for designing future nero-symbolic architectures. Because the utility of HRRs for cognitive-science tasks are already well established, our application area will focus on a multi-label machine learning task to show how HRRs can provide potential value to research in areas it has not been previously associated with.

Backpropagating through HRRs is ineffective if done naively, and our goal is to show how to rectify this issue with a simple projection step and build viable loss functions out of HRR operations. 
We choose to do this with  eXtreme Multi-Label (XML) classification tasks as one that is impossible for HRRs to tackle today. 
XML tasks have an input $\boldsymbol{x}$, from which we have a large number $L$ of binary prediction problems $\mathcal{Y}_1, \mathcal{Y}_2, \ldots, \mathcal{Y}_L$. 
Our approach will be to represent each of the $L$ classes as a HRR concept vector, and construct an alternative to a fully connected output layer that uses the HRR framework for learning. Our experiments and ablations will focus on the impact of replacing a simple output layer with HRRs and other changes to the HRR approach to better understand its behaviors. State-of-the-art XML performance is not a goal. 

Our contributions and the rest of our paper are organized as follows. In \cref{sec:related_work} we discuss  prior approaches related to our own, and how our work differs from these methods. In \cref{sec:hrr_primer} we  give a brief overview of the HRR operator's details given its niche recognition within the AI/ML community, and introduce a simple improvement that increases the binding stability and effectiveness by $100\times$. Then in \cref{sec:dense_label_rep} we  show how we can leverage HRR's symbolic nature to create a new output layer and loss function combination that provides several benefits to deep extreme multi-label models. With this new method we  demonstrate in \cref{sec:results} that compared to a simple fully-connected output layer we can reduce the output size by as much as $99.55$\%, resulting in a total reduction in  model size by up to $42.09$\%, reduce training time by $41.86$\%, and obtain similar or improved relative accuracy by up to $30\%$. We also show that these results are often competitive with more advanced approaches for XML classification, though do still suffer at the largest output spaces with $670k$ labels. Finally we  conclude in \cref{sec:conclusion}. 

\section{Related Work} \label{sec:related_work}

\citet{SMOLENSKY1990159} elucidated early arguments for neuro-symbolic approaches with the Tensor Product Representation (TPR) that defined the first Vector Symbolic Architecture (VSA), defining binding and unbinding operations that allow symbolic style object manipulation atop some vector field. As argued by \citet{SMOLENSKY1990159} (and echoed by \citet{Greff2020}), symbolic logic has greater power for complex tasks, but is brittle in requiring precise specification, where connectionist deep learning is more robustly able to process the raw inputs to a system, its capacity for more powerful ``logic'' is still limited and hotly debated. Combining these two domains in a neuro-symbolic hybrid system may yield positive results, and several recent works have found success in natural language processing tasks by augmenting recurrent neural networks with TPR inspired designs \cite{pmlr-v139-schlag21a,huang-etal-2018-tensor,NEURIPS2018_a274315e}. However, binding $n$ vectors that are each $d$ dimensions requires $\mathcal{O}(d^n)$ space using TPRs. Thus VSAs that require a fixed dimension are computational preferable, but no work to our knowledge has successfully used a VSA of fixed length in gradient based learning. \citet{Schlegel2020} provide a comprehensive summary and comparison of fixed length VSAs, and the details of how we chose to select HRRs based on their potential for gradient based learning and compute/memory efficacy in current frameworks like PyTorch are presented in \autoref{sec:bind_vsa_select}. Though such fixed-length VSAs have not been used in gradient based learning, they have still proven effective low-power embedded computation enviroments due to low-level hardware optimizations possible with many VSAs \cite{Imani2017,Neubert2016}.

The HRR operation has found significant use in cognitive science to create biologically plausible models of human cognition \cite{Jones2007,Blouw2013,Stewart2014,Blouw2016}. These approaches use the HRR's operations in a primarily symbolic fashion to demonstrate human-like performance on a number of cognitive science tasks \cite{Eliasmith2012}. The vector representation is used as the foundation for biological plausibility after establishing that the operations required to implement HRR are within the plausible sphere of a biological substrate \cite{Singh3667}. When learning has been incorporated in these models it has been primarily through examples of Hebbian updates of spiking models on the outputs of HRR, rather than back-propagation through the HRR operations and preceding inputs/layers \cite{10.3389/fninf.2013.00048}. In our work we  advance the understanding of how to learn through the HRR through propagation in an effective manner and how HRR can be leveraged as an integral part of a neural-symbolic approach.  

Little work exists on gradient based learning with HRR operations explicitly. The only work we are aware that attempts to leverage the symbolic properties of the HRR is by \citet{10.5555/3016100.3016172}, who use the binding operations to connect elements in a knowledge graph as a kind of embedding that combines two vectors of information without expanding the dimension of the representation (e.g., concatenation would combine but double the dimension). More recently \citet{Liao2019} used circular convolution as a direct replacement of normal convolution to reduce model size and inference time, without any leverage of the symbolic properties available within HRRs. While \citet{Danihelka2016} purport to embed the HRR into an LSTM, their approach only augments an LSTM by including complex weights and activations\footnote{HRRs \& circular convolution can naturally exist in the reals. }, and does not actually use HRR as it lacks circular convolution, and all three works do not leverage the inverse operation $\inv$ or seek to leverage any symbolic manipulations. By contrast our work explicitly requires the symbolic manipulation properties of HRR to create a vector-symbolic loss function, and we need to leverage the inverse operator $\inv$, meaning we use the entire HRR framework. We also demonstrate new insights into learning with HRR operations that make them more effective, where prior works have simply used the $\bind$ operator on $\leq 2$ components without further study. 

There has been other work in differentiable neuro-symbolic systems outside of TPRs and VSAs, most notably using first-order logic \cite{Badreddine2020,Mao2019,Serafini2016}. These works represent another powerful alternative approach, though at this time are often more involved in their design and training procedures. We consider these beyond the scope of our current study, where we seek to obtain similar benefits using the simpler HRR that requires only the careful application of FFTs to leverage. This is an easily satisfiable constraint given their heavy optimization and wide spread use amongst existing deep learning frameworks. 

Our use of extreme multi-label classification problems is due to it being out-of-reach of current HRR methods, which we found produced random-guessing performance in all cases. There exists a rich literature of XML methods that tackle the large output space from the perspective of decision trees/ensembles \cite{pmlr-v28-bi13,10.1145/2939672.2939756,10.1145/2623330.2623651,pmlr-v28-weston13,pmlr-v80-siblini18a}, label embedding regression \cite{Jalan2019,10.1145/3097983.3097987,NIPS2015_5969,10.5555/3044805.3044873}, naive bayes \cite{niculescu-mizil17a}, and linear classifiers \cite{10.1145/3289600.3290979,10.1145/3018661.3018741}. There also exist deep learning XML methods that use either a fully-connected output layer \cite{10.1145/3077136.3080834} and others that use a variety of alternative approaches to dealing with the large output space \cite{Jiang2021,10.1145/3437963.3441810,10.1145/3437963.3441807,10.1145/3394486.3403368,NEURIPS2019_69cd21a0,NIPS2019_8817,NEURIPS2019_78f7d96e,Wydmuch2018}. The variety of approaches is orthogonal to the purpose of this work, which is to investigate HRRs. We will leverage only some of these works to modify their architectures to use a fully-connected output layer (if they do not already) and to use our HRR approach, so that we can observe the impact of the HRR specifically on the model. In our experimentation we found that the mechanisms for handling the high cardinality of outputs is highly architecture tuned and specific, making it difficult to do broader comparisons. Our work will explore fully-connected, convolutional, and recurrent architectures to show that our HRR approach can work broadly across many types of architectures that have been used, but it is not our goal to be the ``best'' at XML by any metric. The XML task's value to our study is a problem that current HRR methods could not tackle, and where we can apply the symbolic properties of HRR in developing the solution in a way that demonstrates how one may design HRR based networks for other tasks.

\section{Holographic Reduced Representation} \label{sec:hrr_primer}

The original paper by \citet{Plate1995} introduced the HRR approach and developed significant theory for its use as a symbolic approach to AI on a neural substrate. 
We will quickly review its foundation and numerical issues that lead to our projection based improvement.
The HRR starts with circular convolution as a "binding" operator. We use Plate's notation of $\bind$, and a definition of circular convolution using an FFT $\mathcal{F}$ and inverse FFT ($\mathcal{F}^{-1}$) is given in \cref{eq:hrr_bind}. 
It is easy to see that the binding operator of circular convolution $\bind$ is commutative. What is particularly interesting is the ability to \textit{unbind} terms that have been bound by this operator. Let  $\mathcal{F}(\boldsymbol{a})_i$ denote the $i$'th complex feature of the FFT of $\boldsymbol{a}$. By defining an identity function $\mathcal{F}(\boldsymbol{a}^\inv)_i \mathcal{F}(\boldsymbol{a})_i = 1$, we can derive the inverse function \cref{eq:hrr_inv}, where we raise each complex coefficient $z_i$ of the FFT to the $z_i^{-1}$ power before going back to the real domain. 
\noindent\begin{multicols}{2}
\begin{equation}\label{eq:hrr_bind}
    \boldsymbol{a} \bind \boldsymbol{b} = \mathcal{F}^{-1}\left(\mathcal{F}(\textbf{a}) \odot \mathcal{F}(\textbf{b})\right)
\end{equation}
\begin{equation}\label{eq:hrr_inv}
\boldsymbol{a}^\inv = \mathcal{F}^{-1}\left(\frac{1}{\mathcal{F}(\boldsymbol{a})}\right)
\end{equation}
\end{multicols}

Using this, \citet{Plate1995} showed that one can compose sentences like our example $S = \mathit{red} \bind \mathit{cat} + \mathit{blue} \bind \mathit{dog} $ by assigning concepts to vectors in a $d$ dimensional space. We can unbind components with the inverse operation, giving $S \bind \mathit{cat}^\inv \approx \mathit{red}$, or we can check for the existence of a term by checking that $S^\top \mathit{cat}^\inv \approx 1$. A term not present, like "cow", would behave as $S^\top \mathit{cow}^\inv \approx 0$. 
These sentences can be constructed with more complex logic. The binding is also distributive (and associative), so $\boldsymbol{a} \bind (\boldsymbol{x} + \boldsymbol{y}) = \boldsymbol{a} \bind \boldsymbol{x} + \boldsymbol{a} \bind \boldsymbol{y}$. This allows us to construct complex symbolic relationships. As an example, we could have: $S = \mathit{first} \bind (\mathit{name} \bind \mathit{Jane} + \mathit{age} \bind 30) + \mathit{second} \bind (\mathit{name} \bind \mathit{Joe} + \mathit{age} \bind 33)$, and to determine the name of the first person we can perform $\mathit{first}^\inv \bind \mathit{name}^\inv \bind S \approx \mathit{Jane}$.

To make this symbolic manipulations effective, \citet{Plate1995} introduced two points of sufficiency. First, a vector $\boldsymbol{a}$ used for HRR operations should have its values sampled from a normal distribution: $\boldsymbol{a} \sim \mathcal{N}\left(0, \mathrm{I}_d \cdot d^{-1}\right)$. 
This allows binding and unbinding to work in expectation. 
Second, 
is that the
inverse $\inv$ is numerically unstable. The remedy developed is the \textit{pseudo} inverse $\invA$, given by 
$\boldsymbol{a}^{\invA} =\left[a_{1}, a_{d}, a_{d-1}, \ldots, a_{2}\right]$.
This approximation was justified by noting that in polar form one obtains $\mathcal{F}_{j}\left(\mathbf{a}^{\inv}\right)=\frac{1}{r_{j}} e^{-i \theta_{j}}$ and
$\mathcal{F}_{j}\left(\mathbf{a}^{\invA}\right)=r_{j} e^{-i \theta_j}$, which are off by the reciprocal of the complex magnitude. Plate and others \cite{Eliasmith2012} use the pseudo inverse $\invA$ exclusively 
as
the error due to approximation of $\inv$ was smaller than the errors caused by numerical instability. In \autoref{sec:hrr_worked} we provide additional illustration about the potential modeling with HRRs, and worked examples demonstrating why the math works is provided in \autoref{sec:hrr_worked}.

\subsection{Improved HRR Learning} \label{sec:improved_projection}

The only prior work backpropogating through $\bind$ only needed to bind two items together at a time \cite{10.5555/3016100.3016172}, where we need to bind and represent tens-to-hundreds of thousands of concepts.  Prior work also made no use of the unbinding with $\inv$, which we require. 
We found that the default method proposed by \citet{Plate1995} was not sufficient for our work, and needed improvement. In particular, we needed better detection if a vector $\boldsymbol{y}$ (read, \textit{label}) has been bound to a statement $S$ (read, \textit{output}). 
We achieve this by defining a new \textit{complex unit magnitude projection}, which is given by \cref{eq:complex_projection}.

\begin{wrapfigure}[4]{r}{0.4\columnwidth}
\begin{equation}\label{eq:complex_projection}
    \pi (\boldsymbol{x}) = \mathcal{F}^{-1} \left(\ldots,  \frac{\mathcal{F}(\boldsymbol{x})_j }{ \left|\mathcal{F}(\boldsymbol{x})_j\right| }, \ldots \right)
\end{equation}
\end{wrapfigure}
We change the initialization to
$\boldsymbol{a} \sim \pi\left(\mathcal{N}\left(0, \mathrm{I}_d\cdot d^{-1}\right)\right)$ which ensures the vectors we use for HRR operations are unitary, meaning the complex magnitude is one. In this case we obtain $\boldsymbol{a}^\inv = \boldsymbol{a}^\invA$ because in polar form we have $r_j = 1$ $\forall j$, giving $\mathcal{F}_{j}\left(\mathbf{a}^{\inv}\right)=\frac{1}{1} e^{-i \theta_{j}} = \mathcal{F}_{j}\left(\mathbf{a}^{\invA}\right)=1 e^{-i \theta j}$. 

\begin{wrapfigure}[24]{r}{0.5\columnwidth}
\vspace{-10pt}
    \centering
    \adjustbox{max width=0.5\columnwidth}{
        \input{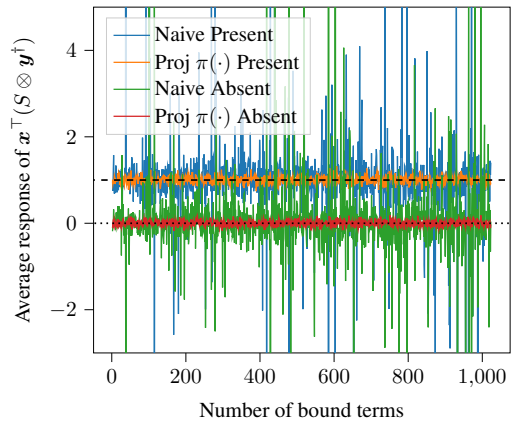}
    }
    \caption{
    Demonstration of the %
    HRR query checking if 
    $\boldsymbol{x} \bind \boldsymbol{y}$ 
    is in the HRR vector $S$. If present, the response value (y-axis) should return $\approx$1.0 (dashed line). If absent, the response should $\approx 0$ (dotted line). 
    Naive use of HRR ({\color{blue} blue} \& {\color{OliveGreen} green}) results in more noise
    with present and absent results both returning values far above and below the expected range $[0, 1]$. Our projection 
$\pi$ ({\color{orange} orange} \& {\color{red} red}) keeps values stable and near their expected values.   
}
    \label{fig:hrr_complex_proj_impact}
\end{wrapfigure}
This has a number of benefits. First, we make the numerically stable $\invA$ have a mathematically equivalent result to the true inverse, removing that source of error. Second, the $\invA$ is faster to calculate requiring a simple shift of values that takes $\bigO(d)$ time to perform, compared to $\bigO(d \log d)$ for the FFTs in $\inv$. Third, we find that this complex projection significantly increases the number of concepts that can be bound together and the accuracy of retrieving them. Using standard HRR initialization we found it difficult to use more than 10 bindings in a statement, but our simple change allows us to improve query accuracy by up to $100\times$ as many bound vectors.

To  demonstrate the effectiveness of our projection step, we create a statement vector $S = \sum_{i=1}^T \boldsymbol{a}_i \bind \boldsymbol{b}_i$. We can ask if two vectors $\boldsymbol{x} \bind \boldsymbol{y} \in S$ by checking that $\boldsymbol{x}^\top (S \bind \boldsymbol{y}^\inv) \approx 1$. If the pair are not bound in $S$, then $\boldsymbol{x}^\top (S \bind \boldsymbol{y}^\inv) \approx 0$ meaning $\boldsymbol{x} \bind \boldsymbol{y} \notin S$. We explicitly use this problem formulation because it is integral to our approach to XML. This is shown for a 256 dimension space in \cref{fig:hrr_complex_proj_impact} where the black dashed line marks the "in" case and the dotted black line the "not in" case. Notice that the naive HRR approach has present values that look absent, and absent values that look present, with unexpectedly large and small magnitudes. The variance of naive HRR also increases as the number of terms bound together increases. In contrast, our projection $\pi$ keeps the response around 1 and 0 for present and absent tokens respectively, even when binding 1024 vectors in only 256 dimensions. This is critical for us to represent large output spaces in a smaller source dimension, and removes the need for complex averaging and associative cleanup approaches attempt by  \citet{Danihelka2016}. Plate's original theory indicated binding capacity would increase linearly with dimension size $d$, which we found did not hold for naive HRRs, but does for our projected variant. Additional details and experiments on binding capacity, and selection of HRRs as opposed to alternative binding operations, is provided in \autoref{sec:bind_vsa_select}. 

\section{Dense Label Representations} %
\label{sec:dense_label_rep}

Now we will define our new neuro-symbolic approach to XML tasks by leveraging the HRR operations with our improved initialization from \cref{sec:improved_projection}. We define $L$ as the number of symbols (or labels) in the given task dataset. We will use a $d' \ll L$ dimensional output space $\mathbb{R}^{d'}$, which is a hyper-parameter we can choose.  We will define two vectors $\boldsymbol{p}, \boldsymbol{m} \in \mathbb{R}^d$ to represent the concepts of a class being \textit{present} and \textit{missing} (or "negative" / "absent") from the label space $\mathcal{Y}_{1, \ldots, L}$. We initialize $\boldsymbol{p} \sim \pi(\mathcal{N}(0, \mathrm{I}_d\cdot d'^{-1}))$, and then select $m$ to be any vector orthogonal to $\boldsymbol{p}$.

We now define some additional notation and terms to describe our approach. We will use $\mathcal{Y}^p$ to denote the set of ground truth labels present for a given datapoint $\boldsymbol{x}$ and $\mathcal{Y}^m$ for the missing labels, where $\mathcal{Y}^p \cup \mathcal{Y}^m = \mathcal{Y}_{1, 2, \ldots, L}$. For every individual classification problem $\mathcal{Y}_i$, we will use $\boldsymbol{c}_i \sim \pi(\mathcal{N}(0, \mathrm{I}_d\cdot d'^{-1}))$ to represent each class with a vector. The vectors $\boldsymbol{p}, \boldsymbol{m}, \boldsymbol{c}_{1, 2, \ldots, L}$ will all be initialized as we have described, and will not be altered during training. To denote the neural network we will train  and its output, we will use $f(\boldsymbol{x}) = \hat{\boldsymbol{s}} \in \mathbb{R}^{d'}$. 

\textbf{HRR Label Vector:}
We begin by converting the labels for a data point $\boldsymbol{x}$ into a \textit{labels vector} $\boldsymbol{s} \in \mathbb{R}^{d'}$. This vector $\boldsymbol{s}$ will be constructed using the HRR framework to be a neuro-symbolic representation of the original labels, and is defined by \cref{eq:hrr_label}. 
Thanks to the the commutative property of $\bind$, the "present" vector $\boldsymbol{p}$ can move outside the summation and be bound with all present classes, and the non-present vector $\boldsymbol{m}$ can be bound with every item not present. This will create a statement $\boldsymbol{s}$ with technically hundreds of thousands of bound components that we could query. 
We can compute this efficiently in $\bigO(|\mathcal{Y}^p|)$ 
by computing once 
the "all labels" vector $\mathcal{A} = \sum_{i=1}^L \boldsymbol{c}_i$. This is done once at initialization, and then we can re-write \cref{eq:hrr_label} into the more efficient form of \cref{eq:hrr_label_fast}, leveraging the symbolic properties of HRR to avoid computation. 
\noindent\begin{multicols}{2}\noindent
\begin{equation} \label{eq:hrr_label}
\mathbf{s}= \overbrace{\sum_{\boldsymbol{c}_p \in \mathcal{Y}^p} \mathbf{p} \bind \boldsymbol{c}_p}^{\text{Labels present}} + 
\overbrace{\sum_{\boldsymbol{c}_m \in \mathcal{Y}^m} \boldsymbol{m} \bind \mathbf{c}_{m}}^{\text{Labels absent}}
\end{equation}
\begin{equation} \label{eq:hrr_label_fast}\noindent
=
 \mathbf{p} \bind \sum_{\boldsymbol{c}_p \in \mathcal{Y}^p} \boldsymbol{c}_p + 
\boldsymbol{m} \bind \left(\mathcal{A} - \sum_{\boldsymbol{c}_p \in \mathcal{Y}^p} \boldsymbol{c}_p \right)
\end{equation}
\end{multicols}

\textbf{HRR XML Loss:}
We now have the output of our network $\hat{\boldsymbol{s}}$ and a target vector $\boldsymbol{s}$ that exist in a $d' \ll L$ dimensional space. A straightforward idea would be to take an embedding style approach to make the loss function $\ell(\boldsymbol{s}, \hat{\boldsymbol{s}}) = \|\boldsymbol{s}-\hat{\boldsymbol{s}}\|_2^2$. This is intuitive and matches prior approaches to XML \cite{NIPS2015_5969,Jalan2019}. However, we found that such an approach resulted in no learning and degenerate random-guessing performance. This is because the regression formulation targets the overall errors in exact values, rather than the real goal of being able to extract the present classes from $\hat{\boldsymbol{s}}$. Because binding many values in a fixed dimension $d'$ introduces noise, the regression approach attempts to learn this underlying noise that is not meaningful to the actual problem. 

Our solution to this issue is to again leverage the properties of HRR operations to create a loss based on our ability to accurately query the predicted statement $\boldsymbol{s}$ for the same components we would expect to find in the true label $\boldsymbol{s}$. We remind the reader that we can query if a class $\boldsymbol{c}_p \in \boldsymbol{s}$ by checking if $\boldsymbol{s}^\top \boldsymbol{c}_p^\invA \approx 1$. We normalize this dot product, i.e., use the cosine similarity $\cos(\boldsymbol{s}, \boldsymbol{c}_p^\invA) = \frac{\mathbf{s}^\top  \mathbf{c}_p^\invA}{\|\mathbf{s}\|\|\mathbf{c}_p^\invA\|}$  to perform these queries in a manner that limits the response magnitude to $[-1, 1]$, which will prevent degenerate solutions that attempt to maximize magnitude of response over query efficacy. 

With the bounded cosine response, our loss has two components. First we build a loss term that
goes
though all present labels $\mathcal{Y}^p$, checking that each
can be extracted. This involves computing $\mathbf{p}^\invA \bind S$ to extract the symbolic statement representing all present vectors (i.e., $\mathbf{p}^\invA \bind S \approx \sum_{\boldsymbol{c}_p \in \mathcal{Y}^p} \boldsymbol{c}_p$). Then we use each label HRR vector $\boldsymbol{c}_p$ and check if it is present in the extracted output. This results in \cref{eq:hrr_loss_pos}

Next we 
confirm that the absent labels 
(the vast majority)
are \textit{not} queryable from the output. Rather than enumerate all $L - |\mathcal{Y}^p|$ labels, or perform lossy negative sampling, we can instead leverage the symbolic properties of HRR. We will compute $\boldsymbol{m}^\invA \bind \boldsymbol{s}$ to extract the representation of all non-present classes, and perform a dot product against all present classes that are expected. This gives 
us
\cref{eq:hrr_loss_neg}.

\noindent\begin{multicols}{2}\noindent
\begin{equation}\label{eq:hrr_loss_pos}
    J_p = \sum_{\boldsymbol{c}_p \in \mathcal{Y}^p} (1 -  \cos{(\mathbf{p}^{\invA} \bind \hat{\mathbf{s}} , \boldsymbol{c}_p}))
\end{equation}
\begin{equation}\label{eq:hrr_loss_neg}
    J_n = \cos{\left(\mathbf{m^{\invA}} \bind \hat{\mathbf{s}}, \sum_{\boldsymbol{c}_p \in \mathcal{Y}^p} \boldsymbol{c}_p \right)}
\end{equation}
\end{multicols}

This negative component of the loss works because it is minimized by having the absent labels $\boldsymbol{m}^\invA \bind \boldsymbol{s}$ and present labels $\sum_{\boldsymbol{c}_p \in \mathcal{Y}^p} \boldsymbol{c}_p$ be disjoint (i.e., no overlap). If there is any overlap, the dot product between these terms will increase, and thus, increase the error. 

Through this neuro-symbolic manipulation we can create a loss term that simultaneously considers the presence/absence of all $L$ labels in only $d' \ll L$ dimensions and $\bigO(|\mathcal{Y}^p|)$ time. 
The final loss is simply $\ell(\boldsymbol{s}, \hat{\boldsymbol{s}}) = J_p + J_n$. 

We remind the reader that the target label $\boldsymbol{s}$ is present in the form of knowing the exact $\boldsymbol{c}_p \in \mathcal{Y}^p$ vectors to use. One could, if  preferred, denote this loss as $\ell(\mathcal{Y}^p, \hat{\boldsymbol{s}})$.  We also note that expectation may be that a new hyper-parameter $\lambda$ is needed to balance between the relative importance of \cref{eq:hrr_loss_pos} and \cref{eq:hrr_loss_neg}. However, we find that no such parameter is needed and simply adding them together is effective.

\section{Experiments \& Analysis} \label{sec:results}

We will now conduct an experimental evaluation of our new HRR based approach to XML classification. We emphasize to the reader that this is to show that the HRR approach of symbolically modeling a goal, and then back-propagating through the vector-symbolic HRR, works and has potential utility broadly. We do not argue for "dominance" at the XML task or to supplant all current approaches. XML is simply a task we are applying the HRR to show the pros (and some cons) of the HRR approach. To this end we will show results with a Fully Connected (FC),  Convolutional Neural Network (CNN) based network, and attention augmented Long-Shot Term Memory (LSTM) networks. 
The CNN and LSTM models are applied to every dataset for which the original text was available, only FC is applied to every dataset since fixed-length feature vectors were always provided\footnote{We did contact \cite{Bhatia16} for all text data but received no reply.}. 
Each will be endowed with our HRR dense label representations  as described in \cref{sec:dense_label_rep} using our improved HRR initialization from \cref{sec:hrr_primer} (becoming HRR-FC, HRR-CNN, and HRR-LSTM). The CNN \& LSTM are taken from the XML-CNN~\cite{10.1145/3077136.3080834} and AttentionXML~\cite{NIPS2019_8817} works. The AttentionXML's hierarchical prediction head is replaced with a fully connected layer in our experiments because we are interested in studying the impact of FC $\to$ HRR replacement, and how the change to an HRR impacts the standard network behavior. Using the XML tasks we will study the impact of the HRR an ablate choices like the use of fixed vs learned values for the $\boldsymbol{p}$ \& $\boldsymbol{m}$ vectors, as well as the label vectors $\boldsymbol{c}_p$, impact of increasing the HRR dimension $d'$, and impact on model size and training time. We also ablated the impact of our projection step $\pi(\cdot)$, and found that in \textit{all} cases skipping the projection step caused degradation to random-guessing performance, and thus will not be explicitly included in results.

\subsection{Datasets \& Evaluation Metrics}
To assess the impact of dense label representations, we use eight of the datasets from \citet{Bhatia16}. Table \ref{tab:exp_extreme_label_dataset_stats} (see appendix) provides statistics about each dataset used for experiments. The average number of samples per label varies from 2.29-1902.15, and the average number of labels per data point varies from 2.40-75.54. \citet{Bhatia16} split the dataset into small scale and large datasets depending on the number of labels in each input sample. Small scale datasets consist of at most $5000$ labels. 
The features are a bag-of-words representation of the text in every dataset. Negative labels are as much as 124321$\times$ more populous than positive labels per point. TF-IDF style pre-processed features are available for all datasets, but the original text is not. We were unable to obtain the original text for all datasets, and models which require the original text (CNN \& LSTM) thus are not tested when the raw text is not available.

We will consider two primary metrics in our evaluation that are common in XML work, Precision at $k$ (P@$k$) and the propensity score at $k$ (PSP@$k$).  Given $\text{rank}_k(\hat{\mathbf y})$ as the rank of all the labels in $\hat{\mathbf y}$ and $p_l$ is the relative frequency of the $l$'th label, the P@$k$ (\autoref{eq:p_k}) measures raw predictive accuracy of the top-$k$ guesses, and PSP@$k$ (\autoref{eq:exp_prop_precision_k}) down-weights correctly predicting frequent classes to counteract the power-law distribution of labels common in XML problems. 

\noindent\begin{multicols}{2}\noindent
\begin{equation}\label{eq:p_k}
    \text{P}@k := \frac{1}{k} \sum_{l\in \text{rank}_k (\hat{\mathbf y})} \mathbf y_l
\end{equation}
\begin{equation}
\text{PSP}@k := \frac{1}{k} \sum_{l\in \text{rank}_k (\hat{\mathbf y})} \frac{\mathbf y_l}{p_l}
\label{eq:exp_prop_precision_k}
\end{equation}
\end{multicols}

For brevity we will focus on $k=1$ in most experiments, but found that $k \in [1,5]$ scores were all highly correlated and did not meaningfully alter any results. Additional metrics we considered are mentioned in \autoref{eq:additional_metrics}, but we found them so highly correlated with either P@$k$ or PSP@$k$ as to make them redundant.

\subsection{Network Architectures} \label{sec:archs}

The baseline multi-label network is a fully-connected (FC) network with two hidden layers. Both hidden layers have the same size of $512$ with a $\textsf{ReLU}$ activation. The basline network has $L$ outputs trained with binary cross entropy (BCE) loss with appropriate sigmoid activation. 
Our HRR version of this network (HRR-FC)  uses the same architecture for input and hidden layers. But unlike the multi-label network, the output layer is significantly constrained (size is $512 \times$ $d'$) where $d'$ is the size of dense label representation (\cref{sec:dense_label_rep} for more details).
We note that this gives our HRR Network less parameters to solve the same problem, giving it a  disadvantage in capacity. 

For a CNN based model we will use the XML-CNN approach of \cite{10.1145/3077136.3080834}. Their original architecture with their code is used as the baseline, and our modified version (HRR-CNN) has the same architecture but replaces the output layer with the HRR approach as we used in HRR-FC just described. We note that the original code selected best results from the test set, leading to over-fitting. We have corrected this issue which prevents us from achieving the same results previously published. 

For the LSTM model, we use the attention based bidirectional approach of AttentionXML~\cite{NIPS2019_8817}. The approach used for the output prediction of AttentionXML is involved in a hierarchical sequence of fully connected layers applied per output token to produce a single scalar output, which makes it non-viable to directly convert to an HRR based approach. For this reason we replace the final hierarchical prediction heads of AttentionXML with the same attention mechanism but use a standard fully-connected output layer like the FC and XML-CNN models do, and denote this as the "LSTM" model in results. Our version of AttentionXML with the dense HRR label approach is denoted as HRR-LSTM. 

Our goal is not to determine the most accurate possible XML model, and for this reason we do not perform any expensive hyper-parameter search over the architectures for any of the six models considered (FC, CNN, LSTM, and HRR variants). Doing so to maximize model accuracy would require considering many parameters and source of variation to make a robust conclusion on accuracy\cite{Bouthillier2021}, but leaves each model on each dataset to have potentially highly different parameters. This obfuscates our true goal\cite{pmlr-v97-bouthillier19a}, which is to understand the impact of the HRR modification in isolation. For this reason we hold as many other variables as constant (depth, layers, neurons per layer, etc) and stick to the defaults found to work well in prior work for non-HRR networks. This allows us to isolate the HRR's impact, and intrinsically puts the HRR network at a disadvantage because it has fewer parameters.

\subsection{XML HRR Accuracy} \label{sec:hrr_accuracy}

We first investigate how the smaller HRR space of $d' \ll L$ dimensions impacts each model's accuracy.  
\autoref{tab:exp_dataset_performance} shows the performance of HRR approach to its respective baselines, evaluated at $k=1$ for brevity.  For all datasets, dimension size $d$ is $400$ except in case of Amazon-13K, Wiki10-31K, Delicious-200K and Amazon-670K where $d$ is $3000$. In many cases our HRR based model approaches the high accuracies of state-of-the-art work, which is informative in that our HRR approach can be implemented with 22 lines of code, compared to hundreds of lines of the more involved methods. We {\ul underline} the datasets that were used in these prior papers and use ${}^a$ to denote being within 5\% (relative) of scores reported by \cite{10.1145/3077136.3080834}, and ${}^b$ for 5\% of the original AttentionXML~\cite{NIPS2019_8817}.  

\begin{table}[!h]
\caption{Accuracy of our baseline models and their HRR counterparts with the same network architecture otherwise. Cases where the HRR outperforms its baseline counterpart are in \textbf{bold}.}
\label{tab:exp_dataset_performance}
\adjustbox{max width=\columnwidth}{
\begin{tabular}{@{}ccccccccc@{}}
\toprule
                     & \multicolumn{2}{c}{Bibtex}         & \multicolumn{2}{c}{Delicious} & \multicolumn{2}{c}{Mediamil} & \multicolumn{2}{c}{{\ul Amazon-12K}} \\ \cmidrule(l){2-3} \cmidrule(l){4-5} \cmidrule(l){6-7} \cmidrule(l){8-9}
Model                & FC               & HRR-FC          & FC            & HRR-FC        & FC       & HRR-FC            & CNN          & HRR-CNN               \\
P@1                  & 46.4             & \textbf{60.3}   & 65.0          & \textbf{66.5} & 84.8     & 83.9              & 89.1         & 84.5                  \\
PSP@1                & 32.5             & \textbf{45.6}   & 64.2          & \textbf{30.0} & 64.2     & 63.7              & 49.2         & 44.2                  \\
\multicolumn{1}{l}{} & \multicolumn{6}{c}{{\ul EURLex-4K}}                                                               & \multicolumn{2}{c}{{\ul Amazon-13K}} \\  \cmidrule(l){2-7} \cmidrule(l){8-9}
Model                & FC               & HRR-FC          & CNN           & HRR-CNN       & LSTM     & HRR-LSTM          & FC           & HRR-FC                \\
P@1                  & 73.4             & \textbf{77.2}${}^{a}$& 47.1     & \textbf{50.0} & 63.0     & \textbf{70.4}     & 93.0${}^{a}$ & \textbf{93.3}${}^{a,b}$\\
PSP@1                & 32.0             & 30.7            & 18.0          & 17.5          & 26.4     & \textbf{26.8}     & 52.6         & 49.6                  \\
\multicolumn{1}{l}{} & \multicolumn{6}{c}{{\ul Wiki10-31K}}                                                              & \multicolumn{2}{c}{{\ul Amazon-13K}} \\ \cmidrule(l){2-7} \cmidrule(l){8-9}
Model                & FC               & HRR-FC          & CNN           & HRR-CNN       & LSTM     & HRR-LSTM          & LSTM         & HRR-LSTM              \\
P@1                  & 80.4${}^{a}$     & \textbf{81.1}${}^{a,b}$& 60.0   & \textbf{74.3} & 83.5     & \textbf{85.0}     & 90.0         & \textbf{93.4}         \\
PSP@1                & 9.46             & 9.19            & 10.4          & 9.88          & 10.6     & 10.5              & 48.7         & \textbf{48.8}         \\
\multicolumn{1}{l}{} & \multicolumn{2}{c}{Delicious-200K} & \multicolumn{4}{c}{{\ul Amazon-670K}}                        &              &                       \\ \cmidrule(lr){2-3} \cmidrule(lr){4-7}
Model                & FC               & HRR-FC          & FC            & HRR-FC        & CNN      & HRR-CNN           &              &                       \\
P@1                  & 21.8             & \textbf{44.9}   & 34.6${}^{a}$  & 19.9          & 14.1     & 6.11              &              &                       \\
PSP@1                & 10.5             & 6.84            & 5.22          & \textbf{8.45} & 9.39     & 1.51              &              &                       \\ \bottomrule
\end{tabular}
}
\end{table}

Each bold case shows HRR improving over its baseline. HRR-FC Bibtex improved $1.30-1.40\times$ and Delicious-200k's P@1 improved $2.06 \times$ despite a $1.52\times$ decrease in PSP@1. HRR-CNN improved the Wiki10-31K results by $1.24\times$. In most cases when the HRR results are not better, they are near the performance of the baseline. Most PSP@1 show a $\leq 1$ point decrease, and relative decrease on P@1 results is minimal. The overall results indicate that the HRR approach is moderately worse for retrieval of low-frequency labels, but often significantly better at high-frequency labels (a trade off that is task specific). The primary outlier in behavior is the Amazon-670K dataset, which appears to be pushing the limits of the HRR approach's ability to distinguish between the 670K label vectors. While this is a negative result in terms of XML, our goal was to improve HRRs: the ability to predict \& retrieve the a correct label 19.9\% of the time out of a space of 670K labels is a near three order of magnitude improvement over the naive HRR without our projection step, which could only perform accurate retrievals of fixed and known vectors in a space of $< 10$ total outputs. 

The high relative performance of our HRR-FC compared to XML-CNN and AttentionXML, and our difficulty replicating XML-CNN when fixing the test-set validation bug, does pose interesting questions about the impact of model choice on XML benchmarks. Similar results have been found 
in systematic review \& replication of work in information retrieval and triplet learning \cite{Dacrema2019,Musgrave2020}, but such an investigation is beyond the scope of our work. 

\subsection{HRR Model Compression \& Runtime}

\begin{wraptable}[11]{r}{0.5\columnwidth}
\vspace{-40pt}
\caption{For each dataset the percentage reduction in parameters of the output layer, and the resulting change for the entire network, by replacing the output layer from a fully-connected softmax with our HRR approach.}
  \label{tab:exp_spn_compression}
    \centering
    \adjustbox{max width=0.5\columnwidth}{
    \begin{tabular}{c | c | c | c}
        \hline
        \multirow{2}{*}{\textbf{Dataset}} & \multirow{2}{*}{\textbf{Dim} $d'$} & \multicolumn{2}{l}{\textbf{\% Compression}}\\\cline{3-4}
        & & \multicolumn{1}{l|}{\textbf{Output}} & \multicolumn{1}{l}{\textbf{Network}}\\
        \hline
        Delicious & 400 & 59.30 & 29.22\\
        EURLex-4K & 400 & 89.98 & 37.80\\
        Wiki10-31K & 3000 & 90.25 & 29.49\\
        Amazon-13K & 3000 & 77.49 & 4.74\\
        Delicious-200K & 3000 & 98.53 & 41.88\\
        Amazon-670K & 3000 & 99.55 & 42.09\\
        \hline
    \end{tabular}
    }
\end{wraptable}
Because of the size of the output space $L$, the output fully connected layer can represent an enormous amount of the weights within a neural network in this problem space. We can see in \autoref{tab:exp_spn_compression} that the HRR approach can reduce the number of parameters in the output layer by 59-99\%, which accounts for 29-42\% of the model's total parameters in most cases. The mild exception is the Amazon-13k corpus, which has a large number of input features and comparatively smaller output space. This shows that our HRR approach naturally provides a more compact representation that could be 
applicable for situations with a high arity of discrete outputs, yet allows learning in a continuous fashion without any sampling tricks.

\label{subsec_hrr_model_runtime}

\begin{wraptable}[15]{r}{0.3\columnwidth}
\vspace{-12pt}
\caption{Speedup in training time for HRR-FC over the FC model. \textbf{Bold} indicates improved runtime.   }
  \label{tab:exp_execution_time}
\begin{tabular}{@{}lr@{}}
\toprule
\multicolumn{1}{c}{Dataset} & \multicolumn{1}{c}{Speedup} \\ \midrule
Bibtex                      & 0.35                        \\
Delicious                   & 0.38                        \\
Mediamill                   & 0.39                        \\
EURLex-4K                   & 0.35                        \\
Wiki10-31K                  & \textbf{1.48}                        \\
Amazon-13K                  & \textbf{1.33}                        \\
Delicious-200K              & \textbf{4.47}                        \\
Amazon-670K                 & \textbf{6.28}                        \\ \bottomrule
\end{tabular}
\end{wraptable}

Fewer parameters leads to a natural speedup in training time too. Table \ref{tab:exp_execution_time} shows the optimization time for a training epoch for a multi-label baseline and HRR-FC where we see an up-to $6.3\times$ reduction in optimization time. The larger the output label space is relative to the rest of the network, the more advantage we obtain. The four cases where HRRs are slower are all the fastest cases, with the smallest dimensions, which prevent the overheads of the HRR implementation from fully realizing their benefit. This is because the cross-entropy calculations are highly optimized, and faster than the cosine similarity and for loops needed by our HRR implementation.

We note that separate from the training time, at inference time on the test set the HRR approach currently takes the same time as the fully-connected output. This is because each positive label $\boldsymbol{c}_p$ is queried against the network's output $\hat{\boldsymbol{s}}$, but this need not be the case. There is further potential for improvement by developing specialized Maximum Inner Product Search (MIPS) kernels that interface nicely with PyTorch, don't require memory copies, and can re-constitute the output labels as needed. These are all engineering challenges we consider beyond scope of the current work, as we are concerned with how to learn through HRRs.

\subsection{Assessing Impact of Hyper-parameters}
By demonstrating non-trivial learning that can match or even improve upon the accuracy of a standard fully-connected output layer that naturally leads to reductions in model size/memory use and improved training time as the models get larger, we have shown that our projection step makes HRRs viable for future learning tasks. This is because without our modification we get consistent random-guessing performance across all datasets. 

However, our larger goal is to gain new information about how to learn with HRRs. As such we now perform a number of ablation experiments to elucidate potential factors relevant to future work outside of XML. This includes the impact of the output dimension $d'$, the effect of increasing model size, and if the HRR vectors should be altered via gradient descent or left static as we have done. Due to space limitations, we will show many results on just the Wiki10-31K dataset in this section as we found it tracks well with overall behavior on other corpora.

\label{subsubsec:exp_label_dim_size}
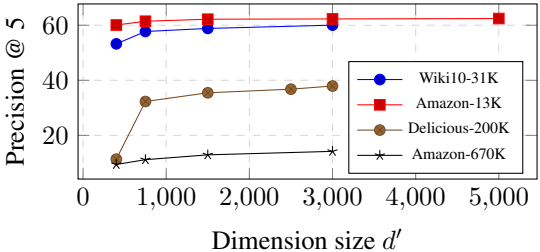
\begin{wrapfigure}[15]{r}{0.55\columnwidth}
\vspace{-12pt}
  \begin{center}
    \begin{tikzpicture}
      \begin{axis}[
          height=0.28\columnwidth,
          width=0.55\columnwidth,          
          grid=major, %
          grid style={dashed,gray!30}, %
          xlabel=Dimension size $d'$, %
          ylabel=Precision @ 5,
        legend columns=1, 
         legend style={font=\tiny},
        legend pos=south east,
        ]
        \addplot table[x=Dimension,y=P5,col sep=tab] {data/wiki10_precision.dat}; 
        \addplot table[x=Dimension,y=P5,col sep=tab] {data/AmazonCat13k_precision.dat}; 
        \addplot table[x=Dimension,y=P5,col sep=tab] {data/DeliciousLarge_precision.dat}; 
        \addplot table[x=Dimension,y=P5,col sep=tab] {data/Amazon670k_precision.dat}; 
        \legend{Wiki10-31K, Amazon-13K, Delicious-200K, Amazon-670K}
      \end{axis}
    \end{tikzpicture}
  \end{center}
\caption{P@5 (y-axis) performance quickly plateaus as the output network size $d'$ (x-axis) increases. The plateau occurs at $d' \leq 10\% \cdot L$, requiring a smaller percentage of all labels $L$ as $L$ increases.}
\label{fig:exp_dim_size_impact}
\end{wrapfigure}
\textbf{Label Dimension Size:}
While training a model with a dense label representation, the vector size $d'$ of the label is fixed. Figure \ref{fig:exp_dim_size_impact} shows the impact of vector dimension size on the model's precision. As observed, Precision@5 increases when the dimension size is increased (and is representative of other metric's behavior). For datasets \textit{Wiki10-31K} and \textit{AmazonCat-13K}, the precision begins to plateau when the dimension size is approximately 10\% of the number of labels in that dataset. The trend is consistent across all Precision@k measurements. For a larger dataset like \textit{Delicious-200K}, the precision levels off at a dimension size of $2500$, which is $1.25$\% of the number of labels. Thus, Figure \ref{fig:exp_dim_size_impact} suggests that HRR can substantially compress the size of the label vector as the number of labels increases (and with it the label distribution itself is increasingly skewed). 

\label{subsubsec:exp_network_layer_size}

\textbf{Network Layer Size:}
Our initial results in \cref{sec:hrr_accuracy} showed that our HRR-FC \& HRR-CNN have comparable or better accuracy, while also having less parameters. We also looked at how these results would change as both models are made larger. Results from Wiki10-31K are 
\begin{wraptable}[15]{r}{0.45\columnwidth}
\centering
\caption{Increasing the number of layers or hidden neurons has less benefit to the FC baseline compared to the HRR approach.}
\label{tab:larger_networks}
\adjustbox{max width=0.45\textwidth}{
\begin{tabular}{@{}ccccr@{}}
\toprule
\textbf{Net.}                   & \textbf{Layers} & \textbf{Hidden} $h$ & \textbf{Out} $d'$ & \multicolumn{1}{c}{\textbf{P@5}} \\ \midrule
\multirow{4}{*}{FC} & 2      & 512        & 30938    & 46.64                   \\
                          & 2      & 2048       & 30938    & 47.91                   \\ \cmidrule(l){2-5} 
                          & 3      & 512        & 30938    & 41.92                   \\
                          & 3      & 2048       & 30938    & 45.25                   \\ \midrule
\multirow{5}{*}{HRR}      & 3      & 512        & 400      & 55.06                   \\
                          & 3      & 2048       & 400      & 55.36                   \\
                          & 3      & 2048       & 750      & 56.84                   \\
                          & 3      & 2048       & 1500     & 57.80                   \\
                          & \textbf{3} & \textbf{2048} & \textbf{3000} & \textbf{58.51}                   \\ \bottomrule
\end{tabular}
}
\end{wraptable}
shown in \cref{tab:larger_networks}. We can see that even moderate changes to the baseline network actually produce a drop in accuracy, which is expected since we tuned the initial architecture to baseline performance. For our HRR-FC more layers or hidden neurons have minor impact, but combined with larger output size $d'$ can increase performance from the initial 53.25\% up to 58.51\%. All of these HRR-FCs still have fewer parameters than the initial baseline network.

\textbf{Updating $\boldsymbol{p}$ \& $\boldsymbol{m}$ Vectors:}
The $\boldsymbol{p}$ and $\boldsymbol{m}$ vectors are the key to the cumulative positive and negative class vectors. It is possible to backpropagate to these vectors and alter them during training, rather than hold them fixed as we have done.  We measure the impact of updating $\boldsymbol{p}$ and $\boldsymbol{m}$ vectors in comparison to keeping them static while the model is being trained.  
We found that  maintaining $\boldsymbol{p}$ and $\boldsymbol{m}$ vectors fixed is beneficial when the dimension size is small ($d' \leq 400$). As  $d'$ increases we observe no impact from  learning the values of $\boldsymbol{p}$ and $\boldsymbol{m}$. This was consistent across all datasets and architectures.

\begin{wraptable}[14]{r}{0.4\columnwidth}
\vspace{-12pt}
\centering
\caption{Ablation test of whether label vectors $\boldsymbol{c}_p$ should be altered during training (right most column) or remain fixed (2nd from right). Results are from Wiki10-31K as a representative dataset. }
\label{tab:fixed_or_grad_labels}
\begin{tabular}{@{}ccrl@{}}
\toprule
           &          & \multicolumn{2}{c}{\textbf{P@5}}                                  \\ \cmidrule(l){3-4} 
\textbf{Hidden} $h$ & \textbf{Out} $d'$ & \multicolumn{1}{c}{\textbf{Fixed}} & \multicolumn{1}{c}{$\nabla_{\boldsymbol{c}_p}$} \\ \midrule
512        & 400      & 55.06                     & 53.50                        \\
2048       & 400      & 55.36                     & 55.45                        \\
2048       & 750      & 56.84                     & 56.75                        \\
2048       & 1500     & 57.80                     & 57.69                        \\
2048       & 3000     & 58.51                     & 58.18                        \\ \bottomrule
\end{tabular}

\end{wraptable}
\textbf{Updating  Label Vectors $\boldsymbol{c}_p$:}
Another question is if we should have allowed the label vectors $\boldsymbol{c}_p$ to be altered during training, rather than holding them fixed at initialized values. We test this in \autoref{tab:fixed_or_grad_labels} using a network with multiple different numbers of hidden neurons $h$ and output dimensions $d'$ for a 2-layer network, where the $\nabla_{\boldsymbol{c}_p}$ is the performance when we allow the concept vectors to be learned. We note that this was implemented by always back-propagating through our complex unit magnitude projection $\pi$, and resulted in no significant performance difference. Experiments without the projection $\pi$ had significantly degraded results that gave random-guessing performance. 

These experiments show that altering the HRR vectors, at least in the XML application, have no  benefit. We had expected this to perform better by giving the network a chance to adjust for any unfavorable initialization. Our current hypothesis is that the network's own parameters are sufficient to learn how to leverage the dynamics of HRR operations, which should still encourage the same properties while training, making the weights of the HRR concept vectors redundant. Further exploration of this is needed in future studies. 

\section{Conclusion} \label{sec:conclusion}

We have improved the initialization of HRRs to increase binding and retrieval accuracy, allowing for a convenient \& differentiable neuro-symbolic approach. To demonstrate potential utility we have applied HRRs to extreme multi-label classification. We observe reduced model size, faster training, and stable-to-improved accuracy. This provides evidence for the general utility and further study of HRRs. The learning dynamics of HRRs are not yet fully understood by our work and have the counter-intuitive behavior that learning with them, but not altering them, tends to have the best results. Compared to prior works looking at HRRs we have significantly improved the viability of building more complex networks that leverage the symbolic manipulation properties of HRRs, which opens the door to a new kind of approach to modeling problems.

\bibliographystyle{IEEEtranN}
\bibliography{refs.bib,references/library.bib,references/datasets.bib}

\clearpage
\appendix
\onecolumn

\section{A Note on Applications and Future Work}

The applications of HRRs may not be immediate, given the approach has been out-of-vogue amongst most machine learning practitioners for many years. Long term we believe improvements in neuro-symbolic learning are important for better generalization of ML methods to novel inputs and situations, as argued by \cite{Greff2020}. In the short term future, we do believe HRRs may have considerable opportunity to provide enhancements. Transformers via their ``query, key, value'' Multi-Headed Attention (MHA) are a natural place to explore HRRs due to the match of logical design, while potentially avoiding MHA's high costs and are supported by similarly motivated analysis by \citet{pmlr-v139-schlag21a} through the lens of an associative memory. Similar recent works on TPR augmented RNNs for natural language processing (NLP) \cite{huang-etal-2018-tensor,NEURIPS2018_a274315e} show value to endowing modern designs with classic symbolic-connectionist ideas. The same inspiration and other neuro-symbolic work on question-answering with TPRs \cite{NEURIPS2018_a274315e} leads us to believe HRRs maybe have similar potential for such systems, and in particular as a way to extract, or augment the knowledge base of an a queryable system in a way that current methods do not yet allow. Broadly we believe many NLP applications of HRRs may exist given the common need to perform binding of subjects to their associated nouns, entity resolution, and the large variety of binding like problems that occur across NLP tasks. Ultimately, we hope that the most interesting work will come from taking new perspectives on how loss functions and problems may be modeled, as we have done in \autoref{sec:dense_label_rep}, to enable new kinds of tasks and applications. 

\section{Understanding Compositional Representations with HRR} \label{sec:hrr_worked}

\begin{figure}[h!]
    \centering
    \adjustbox{max width=\textwidth}{
    \includegraphics{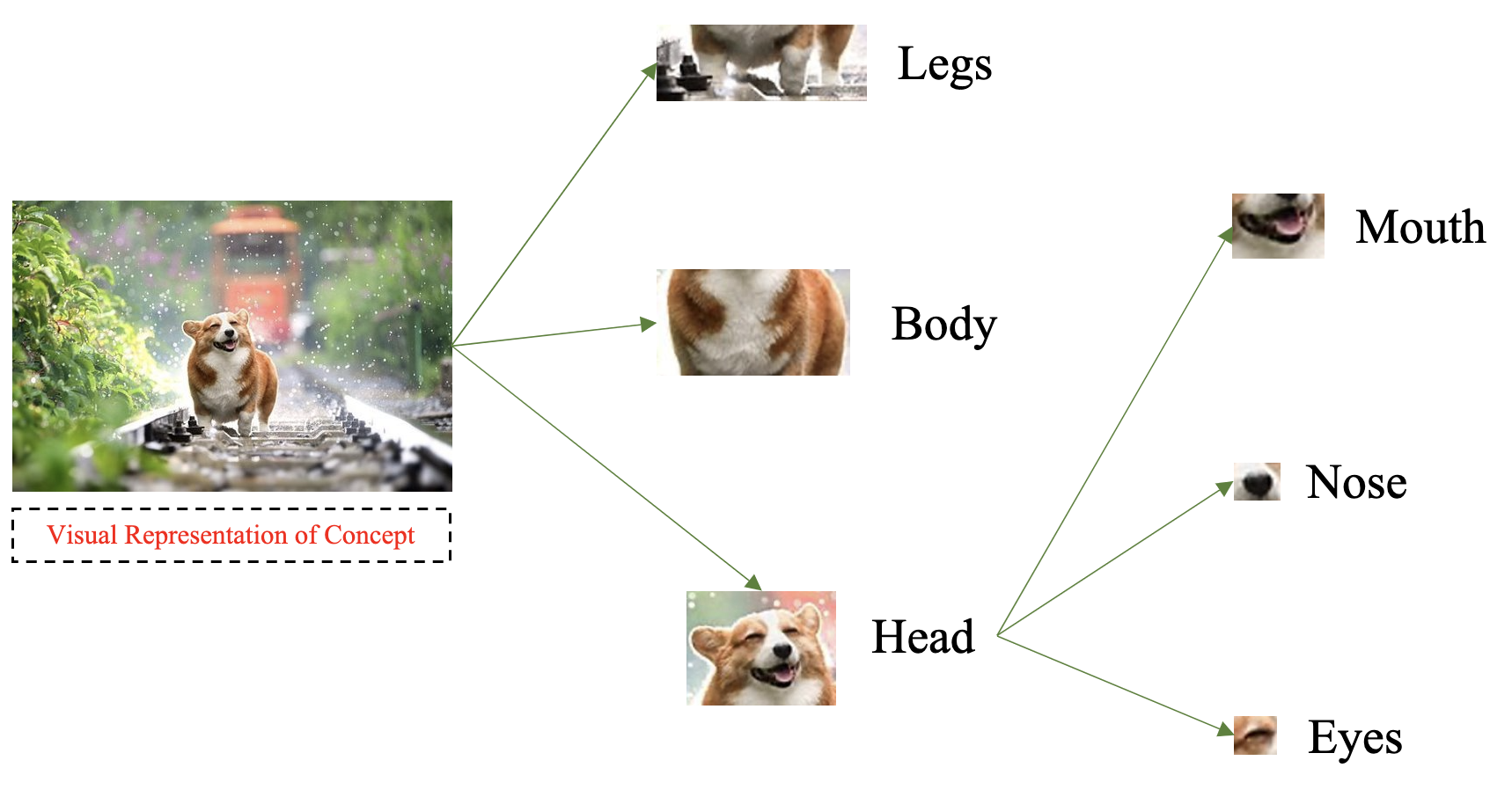}
    }
    \caption{Example representation of \textit{dog} as a combination of different parts. The representation is a two level hierarchy where its \textit{head} can be subdivided into components.}
    \label{fig:hrr_example}
\end{figure}

In this section, we provide an illustrative example of how a compositional representation can be constructed with holographic reduced representations. As shown in \cref{fig:hrr_example}, a dog is represented a combination of the different parts of its body. The representation is in the form of a tree and consists of a two-level hierarchy where the \textit{head} part is further represented as a combination of \textit{eyes}, \textit{nose} and \textit{mouth}. Our objective is to create design a dense vector representation that can represent this hierarchy. There are multiple ways in which a representation can be constructed, such as a binary format, or concatenating individual attribute vectors representations. HRRs allow us construct a dense representation that can be decomposed while maintaining the vector dimension size $d$ constant. 

\begin{figure}[!h]
    \centering
    \adjustbox{max width=\textwidth}{
    \includegraphics{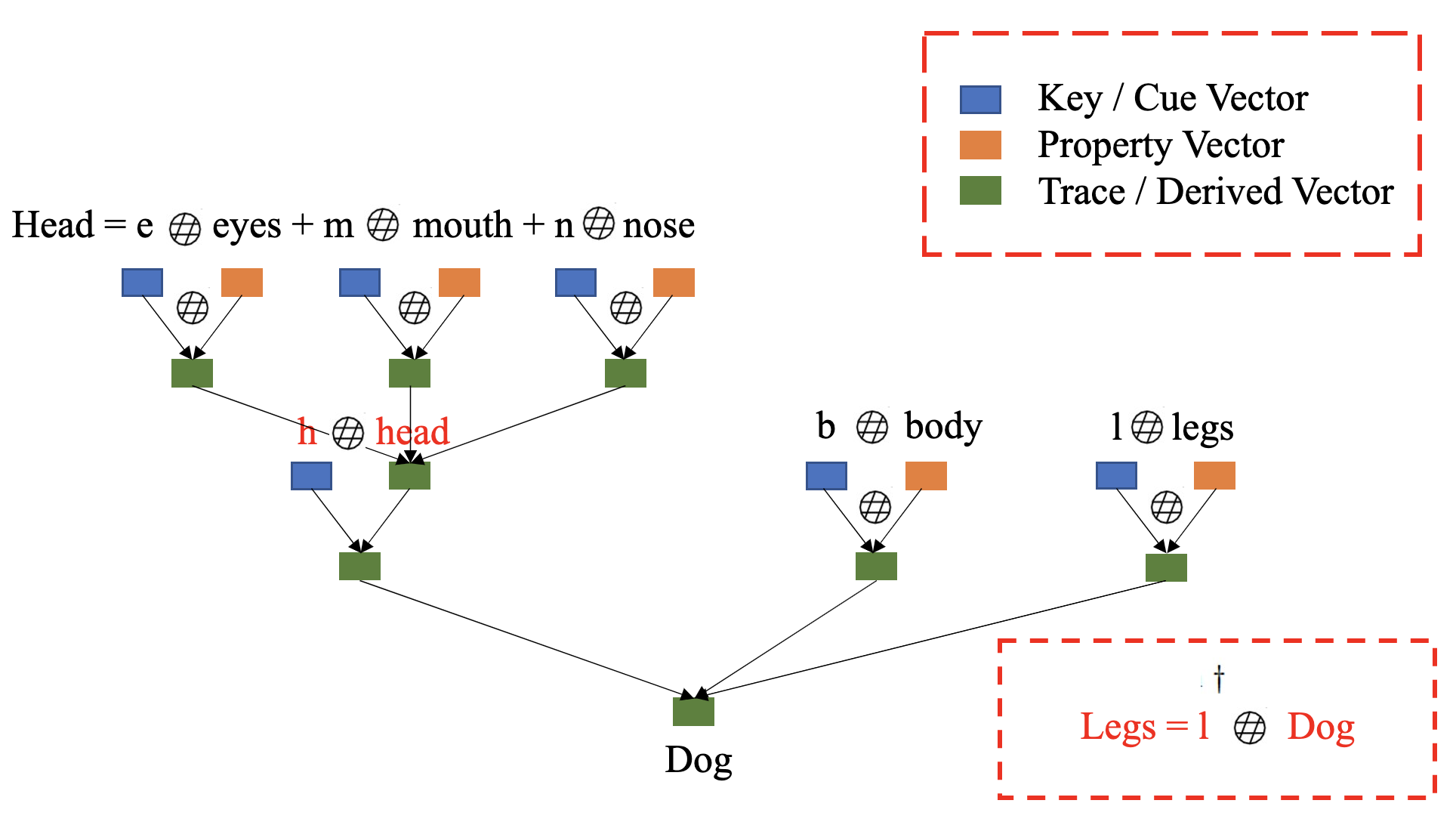}
    }
    \caption{Following is the vector representation of a \textit{dog} using HRR. There are three types of vector representations, namely, the \textbf{Key / Cue} vector, the \textbf{attribute} or \textbf{property} vector and \textbf{trace / derived} vector. Trace vectors can be added to form a combined vector representation. The diagram shows how a two-level hierarchy is represented with HRR. \textbf{It is important to note that the dimension size of the representation remains constant}. To query the vector, the inverse of the key for a given attribute is utilized (with the unbind operator). In this example, we ask the question: \textbf{Does a dog have legs?}}
    \label{fig:hrr_rep_construction_example}
\end{figure}

\cref{fig:hrr_rep_construction_example} shows how HRR can be utilized. As described in \cref{sec:introduction}, each attribute is represented as combination of two vectors: a key (k) and attribute vector. The $\boldsymbol{k}^\inv$ is used to retrieve the original attribute vector.

In the given example, the trace for \textit{dog} (final dense representation) is computed by adding all $\boldsymbol{key} \bind \boldsymbol{attribute}$ pairs. We ask the query: \textbf{Do dogs have legs?} and retrieve the attribute for \textit{legs} by computing $\boldsymbol{l}^\inv \bind \boldsymbol{Dog}$ where $\boldsymbol{l}$ is the key for the attribute \textit{legs} and $\boldsymbol{Dog}$ is the trace vector representing the concept. A simple yes/no response can then be obtained by comparing to the property, via $\boldsymbol{\mathit{legs}}^\top \boldsymbol{l}^\inv \bind \boldsymbol{Dog}$ Because the $\bind$ operation is associative and communicative, we can also ask if dogs have eyes by checking $\boldsymbol{\mathit{eyes}}^\top \boldsymbol{e}^\inv \bind \boldsymbol{Dog}$, though a stronger response will be obtained by using the full structure of the encoding and checking $\boldsymbol{\mathit{eyes}}^\top \boldsymbol{h}^\inv \bind \boldsymbol{e}^\inv \bind \boldsymbol{Dog}$.

The reader may then ask, why should the HRR operation allow us to answer queries like $\boldsymbol{\mathit{legs}}^\top \boldsymbol{l}^\inv \bind \boldsymbol{\mathit{Dog}}$ in a fixed dimensional space? As an example we will reproduce the excellent illustrative worked example by \citet{Plate1995}, followed by a new derivation showing the same nature when distractor terms are included in the statement. 

Consider a $d=3$ dimensional space, where we wish to compute $\boldsymbol{c}^\inv \bind (\boldsymbol{c} \bind \boldsymbol{x})$, we will get the result that:

\begin{equation} 
\boldsymbol{c}^\inv \bind (\boldsymbol{c} \bind \boldsymbol{x}) = 
    \begin{bmatrix}
    \begin{array}{r}x_{0}\left(c_{0}^{2}+c_{1}^{2}+c_{2}^{2}\right)+x_{1} c_{0} c_{2}+x_{2} c_{0} c_{1}+x_{1} c_{0} c_{1} \\ +x_{2} c_{1} c_{2}+x_{1} c_{1} c_{2}+x_{2} c_{0} c_{2}\end{array}\\
    \begin{array}{r}x_{1}\left(c_{0}^{2}+c_{1}^{2}+c_{2}^{2}\right)+x_{0} c_{0} c_{1}+x_{2} c_{0} c_{2}+x_{0} c_{0} c_{2} \\ +x_{2} c_{1} c_{2}+x_{0} c_{1} c_{2}+x_{2} c_{0} c_{1}\end{array} \\
    \begin{array}{r}x_{2}\left(c_{0}^{2}+c_{1}^{2}+c_{2}^{2}\right)+x_{0} c_{0} c_{1}+x_{1} c_{1} c_{2}+x_{0} c_{1} c_{2} \\ +x_{1} c_{0} c_{2}+x_{0} c_{0} c_{2}+x_{1} c_{0} c_{1}\end{array}
    \end{bmatrix} 
\end{equation}

There are two simplifications that can be done to this resulting matrix by exploiting the fact that all elements in the matrix are sampled according to the distribution $x \sim \mathcal{N}(0, \frac{1}{d})$. First there is the pattern $x_i \cdot \prod_{j=0}^{d-1} c_j^2$ on the left hand side. The sum of squared normals will in expectation be equal to 1, but we will subtract that value in our change of variable to create $\xi=\left(c_{0}^{2}+c_{1}^{2}+\cdots+c_{d}^{2}\right)-1$, which will then have the distribution $\xi \sim N\left(0, \frac{2}{n}\right)$. Second, the right-hand side will have $d (d-1)$ products of the form $x_i c_j c_k, \forall j \neq k$. Summing all of these into a new variable $\eta_i$ products $\eta_{i}  \sim \mathcal{N}\left(0, \frac{d-1}{d^2}\right)$. Inserting $\xi_i$ and $\eta_i$ we get:

\begin{equation}
\boldsymbol{c}^\inv \bind (\boldsymbol{c} \bind \boldsymbol{x}) = 
\left[\begin{array}{l}x_{0}(1+\xi)+\eta_{0} \\ x_{1}(1+\xi)+\eta_{1} \\ x_{2}(1+\xi)+\eta_{2}\end{array}\right]=(1+\boldsymbol{\xi}) \tilde{\mathbf{x}}+\tilde{\boldsymbol{\eta}}
\end{equation}

Since both $\xi_i$ and $\eta_i$ have a mean of zero, we get the final result that $\mathbb{E}[\boldsymbol{c}^\inv \bind (\boldsymbol{c} \bind \boldsymbol{x})] = \boldsymbol{x}$, allowing us to recover a noisy approximation of the original bound value. The communicative and associative properties of the HRR's construction then extend this to the more complex statements that are possible, and accumulate the noise of the resulting variables.

To demonstrate this, we will perform another example with $\boldsymbol{c}^\inv \bind (\boldsymbol{c} \bind \boldsymbol{x} + \boldsymbol{a} \bind \boldsymbol{b})$. This will be performed with $d=2$ in order to avoid visual clutter, and results in the equation:

\begin{equation}
\begin{array}{lcc}
\boldsymbol{c}^\inv \bind (\boldsymbol{c} \bind \boldsymbol{x} + \boldsymbol{a} \bind \boldsymbol{b}) & = & 
    \left[\begin{matrix}\frac{c_{0} \left(a_{0} b_{0} + a_{1} b_{1} + c_{0} x_{0} + c_{1} x_{1}\right) - c_{1} \left(a_{0} b_{1} + a_{1} b_{0} + c_{0} x_{1} + c_{1} x_{0}\right)}{\left(c_{0} - c_{1}\right) \left(c_{0} + c_{1}\right)}\\\frac{c_{0} \left(a_{0} b_{1} + a_{1} b_{0} + c_{0} x_{1} + c_{1} x_{0}\right) - c_{1} \left(a_{0} b_{0} + a_{1} b_{1} + c_{0} x_{0} + c_{1} x_{1}\right)}{\left(c_{0} - c_{1}\right) \left(c_{0} + c_{1}\right)}\end{matrix}\right] \\
    & = & 
    \left[\begin{matrix}\frac{\color{red}{a_{0} b_{0} c_{0} - a_{0} b_{1} c_{1} - a_{1} b_{0} c_{1} + a_{1} b_{1} c_{0}} \color{black}{+ c_{0}^{2} x_{0} - c_{1}^{2} x_{0}}}{c_{0}^{2} - c_{1}^{2}}\\\frac{\color{red}{- a_{0} b_{0} c_{1} + a_{0} b_{1} c_{0} + a_{1} b_{0} c_{0} - a_{1} b_{1} c_{1}}\color{black}{ + c_{0}^{2} x_{1} - c_{1}^{2} x_{1}}}{c_{0}^{2} - c_{1}^{2}}\end{matrix}\right]
\end{array}
\end{equation}

Notice that the red highlighted portion of the equation is the product of independent random variables, meaning two important properties will apply: $\mathbb{E}[X Y] = \mathbb{E}[X] \cdot \mathbb{E}[Y]$ and $\operatorname{Var}(X Y)=\left(\sigma_{X}^{2}+\mu_{X}^{2}\right)\left(\sigma_{Y}^{2}+\mu_{Y}^{2}\right)-\mu_{X}^{2} \mu_{Y}^{2}$. Because these random variables have a mean $\mu = 0$, the products result in a new random variable with the same mean and reduced variance as the original independent components. The first property gives 

$$\mathbb{E}[X Y] = \mathbb{E}[X] \cdot \mathbb{E}[Y] = 0 \cdot 0 = 0$$

and the second property gives:

$$\operatorname{Var}(X Y)=\left(\sigma_{X}^{2}+\mu_{X}^{2}\right)\left(\sigma_{Y}^{2}+\mu_{Y}^{2}\right)-\mu_{X}^{2} \mu_{Y}^{2} = \left(\left(\frac{1}{d}\right)^2+0\right) \left(\left(\frac{1}{d}\right)^2+0\right)-0 = \frac{1}{d^4}$$

That reduces each product of $a_i b_j c_k$ into a new random variable with a mean of zero, and then the sum of these random variables, due the the linearity of expectation, will be a new random variable with an expected value of zero. So in expectation, the highlighted red terms will not be present (but their variance due to noise will cause errors, though the variance is harder to quantify due to reuse of random variate across the products). Thus we get the expected result of:

\begin{equation}
    \left[\begin{matrix}\frac{c_{0}^{2} x_{0} - c_{1}^{2} x_{0}}{c_{0}^{2} - c_{1}^{2}}\\\frac{c_{0}^{2} x_{1} - c_{1}^{2} x_{1}}{c_{0}^{2} - c_{1}^{2}}\end{matrix}\right] = \left[\begin{matrix}x_0 \\ x_1\end{matrix}\right]
\end{equation}

Which recovers the original $\boldsymbol{x}$ value that was bound with $\boldsymbol{c}$, even though additional terms (e.g., $\boldsymbol{a}\bind \boldsymbol{b}$ are present in the summation.

\section{Implementation} \label{sec:implementation}

Our implementation for all experiments is included in the appendix, and is based off original XML projects from the authors of AttentionXML and XML-CNN, and as such contain significant code that is specific to the data loaders, their original training pipelines, and other features extraneous to the task of understanding just the code for an HRR. As such we take a moment to demonstrate the PyTorch code one could write (as of 1.8.1 which added revamped support for complex numbers and ffts) to implement our HRR approach. 

First are the operations for binding, the inverse and approximate inverse functions, and our projection step. This can be accomplished in just 10 lines of Python code, as the below block shows. The use of the \textit{real} and \textit{nan\_to\_num} functions are defensive guards against numerical errors accumulating in the fft functions that could cause small complex values to occur in the results of computations. 

\begin{minted}[linenos]{python}
def binding(x, y):
    return torch.real(ifft(torch.multiply(fft(x), fft(y))))
def approx_transpose(x):
    x = torch.flip(x, dims=[-1])
    return torch.roll(x, 1, dims=-1)
unbind = lambda x, y: binding(s, approx_transpose(y))
def projection(x):
    f = torch.abs(fft(x)) + 1e-5
    p = torch.real(ifft(fft(x) / f))
    return torch.nan_to_num(p) #defensive call
\end{minted}

The loss is also easy to implement, and below we show a slice of how most of our models implemented the loss approach. The \texttt{inference} function takes in a \texttt{p\_or\_m} variable that is either the present vector $\boldsymbol{p}$ or the missing vector $\boldsymbol{m}$, extracts the target vector from the prediction (i.e.,
$\mathbf{p}^{\invA} \bind \hat{\mathbf{s}}$
or $\mathbf{m}^{\invA} \bind \hat{\mathbf{s}}$), 
and then L2 normalizes the result so that the down-stream dot product becomes equivalent to the cosine distance. The \texttt{inference} function is then used for computing \texttt{J\_p} and \texttt{J\_n}, but using the abs function instead of the true angular distance as a micro optimization. We obtain the same results regardless of that implementation choice, but the abs calls are just a bit faster to run and avoid add inverse cosine calls. 

\begin{minted}[linenos,breaklines]{python}
def inference(s, batch_size, p_or_m):
    vec = p_or_m.unsqueeze(0).expand(batch_size, self.label_size) #make shapes work
    y = unbind(s, vec) #(batch, dims), extracting the target values from prediction
    y = y / (torch.norm(y, dim=-1, keepdim=True) + 1e-8) #normalize so that results will be cosine scores
    return y

convolve = inference(s, target.size(0), p)
cosine = torch.matmul(pos_classes, convolve.unsqueeze(1).transpose(-1, -2)).squeeze(-1) #compute dot products
J_p = torch.mean(torch.sum(1 - torch.abs(cosine), dim=-1))

convolve = inference(s, target.size(0), m)
cosine = torch.matmul(pos_classes, convolve.unsqueeze(1).transpose(-1, -2)).squeeze(-1)#compute dot products
J_n = torch.mean(torch.sum(torch.abs(cosine), dim=-1))

loss = J_n + J_p # Total Loss.
\end{minted}

As seen in the implementation above, \texttt{J\_p} and \texttt{J\_n} are the positive and negative losses. The \texttt{cosine} value can be positive or negative value ranging from $1$ to $-1$. While inferring if an \texttt{unbind} vector is related to a label vector, we compute the \texttt{cosine} distance. Hence, while computing the loss, we take the absolute value of the cosine in order to maintain the positive loss minimizing towards $0$.   

\section{Binding Capacity and VSA Selection} \label{sec:bind_vsa_select}

HRRs are but one of many possible vector symbolic architectures (VSAs) that one could select. For the purposes of our work, we had four desiderata.

\begin{enumerate}
    \item The VSA vectors should naturally exist in the reals, since most deep learning applications are using real-valued vectors. 
    \item The VSA should be composed entirely of differentiable operations, so that learning may be possible. 
    \item The VSA should be of minimal additional overhead. 
    \item The VSA should be as effective as possible at the binding operation. 
\end{enumerate}

The first two of these items are binary requirements that a VSA either has or not. This excludes many VSAs that operate in the complex domain or discrete spaces, leaving us with three potential candidates: HRRs, continuous Multiply-Add-Permute (MAP-C, distinguishing from its binary alternative)\cite{Gayler1998MultiplicativeBR}\cite{Gosmann:2019:VTB:3334291.3334293}, and the most recently developed Vector-derived Transformation Binding (VTB). Of these three the MAP-C option is least desirable because it requires a clipping operation to project vectors values into the range of $[-1, 1]$, which results in sub-gradients and zero-gradient values that will make optimization more challenging. 

In evaluating the overhead of each method, HRRs and MAP-C are satisficing, they are both composed of operations well defined and optimized by existing deep learning systems. The VTB approach requires a sparse block-diagonal Kronecker product that we found is not well optimized in current tools, often requiring $10\times$ the memory to back-propagate through compared to the HRRs and MAP-C, making it less desirable. We stress we do not believe this to be a fundamental limitation of VTB, but a limitation of current tooling. We are confident a custom implementation will work without memory overheads, but wish to constrain ourselves to already existing functions of PyTorch due to simplicity and expediency.

\subsection{Capacity For Error Free Retrieval}

The last question, VSA effectiveness, then becomes part of the decision process in selecting a final VSA to use. To help elucidate how we came to chose the HRR, we will discuss experimental results on the capacity of the VSAs with respect to problems of the form:

$$S = \sum_{i=1}^n \operatorname{bind}(\boldsymbol{x}_i, \boldsymbol{y}_i)$$

This form of $S$ is the same that we rely on to develop our loss framework in \autoref{sec:dense_label_rep}, and does not capture all the ways that a VSA may be used. This analysis is thus not conclusive to holistic VSA effectiveness,  but it does capture the common form of capacity that we will discuss that influenced our selection. 

To estimate the capacity, after $n$ pairs of items are bound together we attempt to unbind $y_i$ which should return $\operatorname{unbind}(S, \boldsymbol{y}_i) = \hat{\boldsymbol{x}}_i \approx \boldsymbol{x}_i$. There will then be a pool of $n$ random distractor vectors $\boldsymbol{z}_1, \ldots, \boldsymbol{z}_n$, sampled in the same manner used to construct the $\boldsymbol{x}_i$ and $\boldsymbol{y}_i$ values of the VSA being tested. If there exists any $\boldsymbol{z}_j$ such that $\operatorname{cos-sim}(\hat{\boldsymbol{x}}_i, \boldsymbol{x}_i) < \operatorname{cos-sim}(\hat{\boldsymbol{x}}_i, \boldsymbol{z}_j)$, then that $j$'th item is considered to be incorrectly retrieved. So our capacity will be the value of $n$ such that no more than $t$ retrieval errors occur. 

\begin{figure}[!h]
    \centering
    \adjustbox{max width=\textwidth}{
    \begin{tikzpicture}

\definecolor{color0}{rgb}{1,0.647058823529412,0}

\begin{axis}[
legend cell align={left},
legend style={fill opacity=0.8, draw opacity=1, text opacity=1, at={(0.03,0.97)}, anchor=north west, draw=white!80!black},
log basis x={2},
log basis y={2},
tick align=outside,
tick pos=left,
x grid style={white!69.0196078431373!black},
xlabel={Embedding Size $d$},
xmin=18.7246090200019, xmax=10814.6450365765,
xmode=log,
xtick style={color=black},
xtick={4,16,64,256,1024,4096,16384,65536},
xticklabels={\(\displaystyle {2^{2}}\),\(\displaystyle {2^{4}}\),\(\displaystyle {2^{6}}\),\(\displaystyle {2^{8}}\),\(\displaystyle {2^{10}}\),\(\displaystyle {2^{12}}\),\(\displaystyle {2^{14}}\),\(\displaystyle {2^{16}}\)},
y grid style={white!69.0196078431373!black},
ylabel={Capacity $n$},
ymin=6.5921628147024, ymax=466.007907624575,
ymode=log,
ytick style={color=black},
ytick={2,4,8,16,32,64,128,256,512,1024},
yticklabels={\(\displaystyle {2^{1}}\),\(\displaystyle {2^{2}}\),\(\displaystyle {2^{3}}\),\(\displaystyle {2^{4}}\),\(\displaystyle {2^{5}}\),\(\displaystyle {2^{6}}\),\(\displaystyle {2^{7}}\),\(\displaystyle {2^{8}}\),\(\displaystyle {2^{9}}\),\(\displaystyle {2^{10}}\)}
]
\addplot [semithick, red]
table {%
25 8
64 8
121 8
256 8
484 8
1024 12
2025 16
4096 24
8100 24
};
\addlegendentry{HRR}
\addplot [semithick, color0, dashed]
table {%
25 8
64 8
121 16
256 16
484 32
1024 64
2025 128
4096 192
8100 384
};
\addlegendentry{HRR w/ Proj}
\addplot [semithick, blue, dash pattern=on 1pt off 3pt on 3pt off 3pt]
table {%
25 8
64 8
121 8
256 24
484 32
1024 64
2025 96
4096 192
8100 384
};
\addlegendentry{VTB}
\addplot [semithick, green!50!black, dotted]
table {%
25 8
64 8
121 8
256 8
484 24
1024 32
2025 64
4096 96
8100 192
};
\addlegendentry{MAP-C}
\end{axis}

\end{tikzpicture}
    }
    \caption{Capacity based on the ability to represent $n$ items bound together and, compared to $n$ false distractors, correctly identify the true item as the most-similar. As the dimension of the embedding space $d$ increases, most VSAs capacity increases linearly. }
    \label{fig:capacity}
\end{figure}
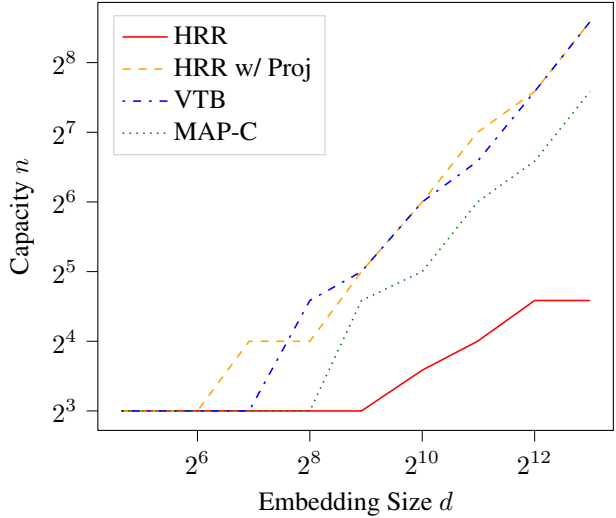

\autoref{fig:capacity} shows the capacity of each method given a threshold of no more than 3\% error, as estimated by 10 trials of randomly selecting all $n$ pairs and distractor items, with $n$ tested at values of $\sqrt{2}^j$. We can see that the naive HRR actually has the worst performance, in part due to its numerical instability/approximation error. It is also important to note that the HRR's original theory developed by \citet{Plate1995} states that the capacity should grow linearly with the dimension size. We find for naive HRRs this is not the case. 

Because HRRs did best satisfying all requirements but the capacity issue, we chose to attempt to improve the HRRs so that they would be more effective\footnote{This work in fact started before the VTB method was published, but was reconsidered when we learned of it.}. As we discussed in \autoref{sec:improved_projection} this can be done with our projection operation, which restores the theoretically expected behavior of linear capacity improvement with dimension size $d$, and brings HRRs to parity with the best performing (in terms of capacity) VSA the VTB. Since the HRR required significantly less memory than VTB, and was slightly faster in our testing, our improved HRR became the most logical choice to move forward with.

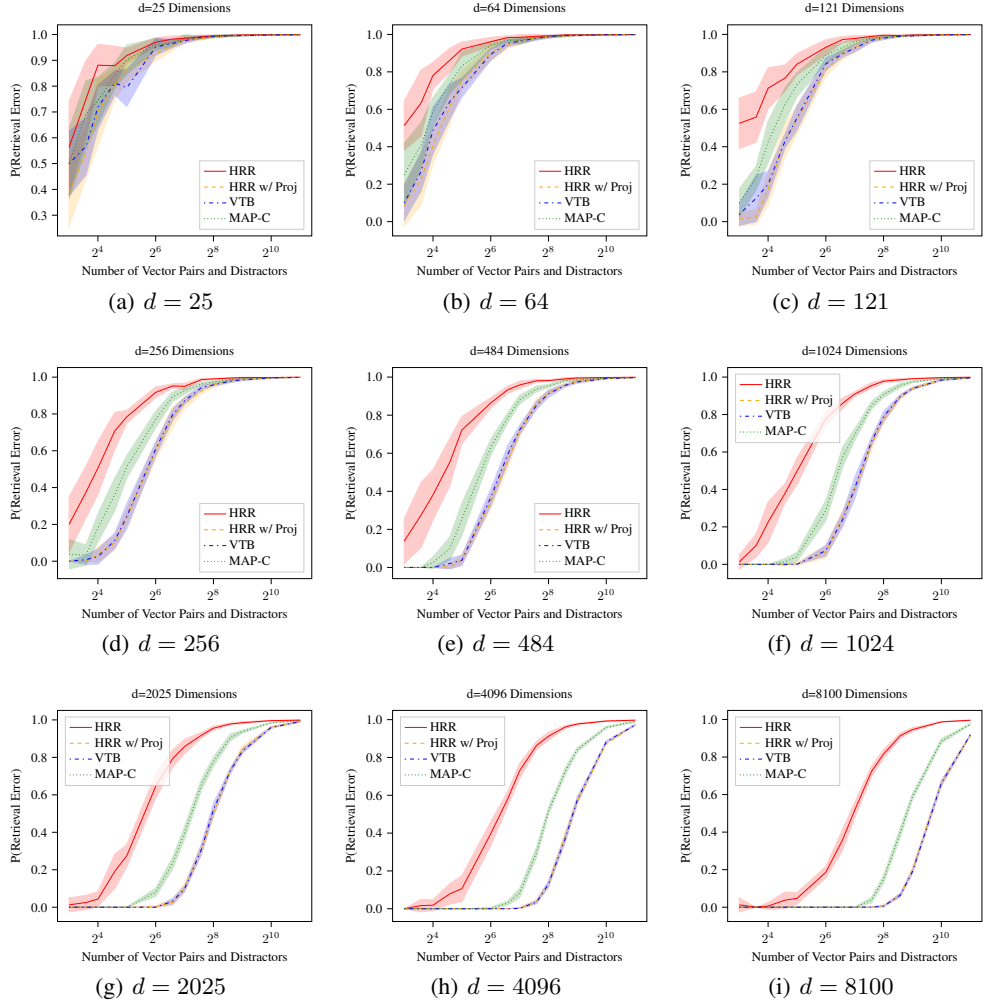
\begin{figure}[!h]
     \centering
     \subfigure[$d=25$]{
         \centering
        \adjustbox{max width=0.3\textwidth}{
        \begin{tikzpicture}

\definecolor{color0}{rgb}{1,0.647058823529412,0}

\begin{axis}[
legend cell align={left},
legend style={fill opacity=0.8, draw opacity=1, text opacity=1, at={(0.97,0.03)}, anchor=south east, draw=white!80!black},
log basis x={2},
tick align=outside,
tick pos=left,
title={d=25 Dimensions},
x grid style={white!69.0196078431373!black},
xlabel={Number of Vector Pairs and Distractors},
xmin=6.06168453445102, xmax=2713.43714878584,
xmode=log,
xtick style={color=black},
xtick={1,4,16,64,256,1024,4096,16384},
xticklabels={\(\displaystyle {2^{0}}\),\(\displaystyle {2^{2}}\),\(\displaystyle {2^{4}}\),\(\displaystyle {2^{6}}\),\(\displaystyle {2^{8}}\),\(\displaystyle {2^{10}}\),\(\displaystyle {2^{12}}\),\(\displaystyle {2^{14}}\)},
y grid style={white!69.0196078431373!black},
ylabel={P(Retrieval Error)},
ymin=0.21982825738121, ymax=1.03752230504418,
ytick style={color=black},
ytick={0.2,0.3,0.4,0.5,0.6,0.7,0.8,0.9,1,1.1},
yticklabels={0.2,0.3,0.4,0.5,0.6,0.7,0.8,0.9,1.0,1.1}
]
\path [draw=red, fill=red, opacity=0.2]
(axis cs:8,0.741472763290954)
--(axis cs:8,0.383527236709046)
--(axis cs:12,0.600928801500014)
--(axis cs:16,0.8)
--(axis cs:24,0.803590178553262)
--(axis cs:32,0.890108901906526)
--(axis cs:64,0.952565911419374)
--(axis cs:96,0.969922560326002)
--(axis cs:128,0.975)
--(axis cs:192,0.984537199197683)
--(axis cs:256,0.991870996220349)
--(axis cs:384,0.99396740849221)
--(axis cs:512,0.996875)
--(axis cs:1024,0.998046875)
--(axis cs:2056,0.999103405586962)
--(axis cs:2056,1.00002110803171)
--(axis cs:2056,1.00002110803171)
--(axis cs:1024,1)
--(axis cs:512,0.999609375)
--(axis cs:384,0.998740924841123)
--(axis cs:256,0.997972753779651)
--(axis cs:192,0.997754467468984)
--(axis cs:128,0.9953125)
--(axis cs:96,0.994660773007331)
--(axis cs:64,0.988059088580626)
--(axis cs:32,0.947391098093474)
--(axis cs:24,0.954743154780071)
--(axis cs:16,0.9625)
--(axis cs:12,0.899071198499986)
--(axis cs:8,0.741472763290954)
--cycle;

\path [draw=color0, fill=color0, opacity=0.2]
(axis cs:8,0.518003831361382)
--(axis cs:8,0.256996168638618)
--(axis cs:12,0.441320045176731)
--(axis cs:16,0.569496168638618)
--(axis cs:24,0.748821429762143)
--(axis cs:32,0.814631809177261)
--(axis cs:64,0.897709085449032)
--(axis cs:96,0.937155934604351)
--(axis cs:128,0.967807378425208)
--(axis cs:192,0.978125)
--(axis cs:256,0.986850818832152)
--(axis cs:384,0.991471946071649)
--(axis cs:512,0.994580865880388)
--(axis cs:1024,0.996369389953246)
--(axis cs:2056,0.998873290302047)
--(axis cs:2056,0.999764841993673)
--(axis cs:2056,0.999764841993673)
--(axis cs:1024,0.999138422546754)
--(axis cs:512,0.997606634119612)
--(axis cs:384,0.998632220595018)
--(axis cs:256,0.995961681167848)
--(axis cs:192,0.991666666666667)
--(axis cs:128,0.986880121574792)
--(axis cs:96,0.975344065395649)
--(axis cs:64,0.955415914550968)
--(axis cs:32,0.904118190822739)
--(axis cs:24,0.867845236904524)
--(axis cs:16,0.830503831361382)
--(axis cs:12,0.642013288156603)
--(axis cs:8,0.518003831361382)
--cycle;

\path [draw=blue, fill=blue, opacity=0.2, line width=0pt]
(axis cs:8,0.625)
--(axis cs:8,0.375)
--(axis cs:12,0.456112506988153)
--(axis cs:16,0.633740809320403)
--(axis cs:24,0.765915250468754)
--(axis cs:32,0.722215480360877)
--(axis cs:64,0.919199507535613)
--(axis cs:96,0.949453984691396)
--(axis cs:128,0.965625)
--(axis cs:192,0.984666507026609)
--(axis cs:256,0.988552100053548)
--(axis cs:384,0.992708333333333)
--(axis cs:512,0.994000430100179)
--(axis cs:1024,0.997443326182861)
--(axis cs:2056,0.998151750972763)
--(axis cs:2056,0.999708171206226)
--(axis cs:2056,0.999708171206226)
--(axis cs:1024,0.999626986317139)
--(axis cs:512,0.997405819899821)
--(axis cs:384,0.998958333333333)
--(axis cs:256,0.995041649946452)
--(axis cs:192,0.995541826306724)
--(axis cs:128,0.9859375)
--(axis cs:96,0.981796015308604)
--(axis cs:64,0.983925492464387)
--(axis cs:32,0.865284519639122)
--(axis cs:24,0.859084749531246)
--(axis cs:16,0.803759190679597)
--(axis cs:12,0.67722082634518)
--(axis cs:8,0.625)
--cycle;

\path [draw=green!50!black, fill=green!50!black, opacity=0.2, line width=0pt]
(axis cs:8,0.636930639376291)
--(axis cs:8,0.363069360623708)
--(axis cs:12,0.562298544197833)
--(axis cs:16,0.653394798157494)
--(axis cs:24,0.793075400876289)
--(axis cs:32,0.846442633914947)
--(axis cs:64,0.940183036160804)
--(axis cs:96,0.954166666666667)
--(axis cs:128,0.979333106213224)
--(axis cs:192,0.980208333333333)
--(axis cs:256,0.987565562669453)
--(axis cs:384,0.990929063287971)
--(axis cs:512,0.994380794997213)
--(axis cs:1024,0.9982421875)
--(axis cs:2056,0.998957183535559)
--(axis cs:2056,0.999972777553935)
--(axis cs:2056,0.999972777553935)
--(axis cs:1024,0.999609375)
--(axis cs:512,0.998587955002787)
--(axis cs:384,0.998654270045362)
--(axis cs:256,0.996809437330547)
--(axis cs:192,0.992708333333333)
--(axis cs:128,1.00035439378678)
--(axis cs:96,0.977083333333333)
--(axis cs:64,0.984816963839196)
--(axis cs:32,0.959807366085053)
--(axis cs:24,0.890257932457044)
--(axis cs:16,0.834105201842506)
--(axis cs:12,0.8210347891355)
--(axis cs:8,0.636930639376291)
--cycle;

\addplot [semithick, red]
table {%
8 0.5625
12 0.75
16 0.88125
24 0.879166666666667
32 0.91875
64 0.9703125
96 0.982291666666667
128 0.98515625
192 0.991145833333333
256 0.994921875
384 0.996354166666667
512 0.9982421875
1024 0.9990234375
2056 0.999562256809339
};
\addlegendentry{HRR}
\addplot [semithick, color0, dashed]
table {%
8 0.3875
12 0.541666666666667
16 0.7
24 0.808333333333333
32 0.859375
64 0.9265625
96 0.95625
128 0.97734375
192 0.984895833333333
256 0.99140625
384 0.995052083333333
512 0.99609375
1024 0.99775390625
2056 0.99931906614786
};
\addlegendentry{HRR w/ Proj}
\addplot [semithick, blue, dash pattern=on 1pt off 3pt on 3pt off 3pt]
table {%
8 0.5
12 0.566666666666667
16 0.71875
24 0.8125
32 0.79375
64 0.9515625
96 0.965625
128 0.97578125
192 0.990104166666667
256 0.991796875
384 0.995833333333333
512 0.995703125
1024 0.99853515625
2056 0.998929961089494
};
\addlegendentry{VTB}
\addplot [semithick, green!50!black, dotted]
table {%
8 0.5
12 0.691666666666667
16 0.74375
24 0.841666666666667
32 0.903125
64 0.9625
96 0.965625
128 0.98984375
192 0.986458333333333
256 0.9921875
384 0.994791666666667
512 0.996484375
1024 0.99892578125
2056 0.999464980544747
};
\addlegendentry{MAP-C}
\end{axis}

\end{tikzpicture}
        }
    }
     \subfigure[$d=64$]{
         \centering
        \adjustbox{max width=0.3\textwidth}{
        \begin{tikzpicture}

\definecolor{color0}{rgb}{1,0.647058823529412,0}

\begin{axis}[
legend cell align={left},
legend style={fill opacity=0.8, draw opacity=1, text opacity=1, at={(0.97,0.03)}, anchor=south east, draw=white!80!black},
log basis x={2},
tick align=outside,
tick pos=left,
title={d=64 Dimensions},
x grid style={white!69.0196078431373!black},
xlabel={Number of Vector Pairs and Distractors},
xmin=6.06168453445102, xmax=2713.43714878584,
xmode=log,
xtick style={color=black},
xtick={1,4,16,64,256,1024,4096,16384},
xticklabels={\(\displaystyle {2^{0}}\),\(\displaystyle {2^{2}}\),\(\displaystyle {2^{4}}\),\(\displaystyle {2^{6}}\),\(\displaystyle {2^{8}}\),\(\displaystyle {2^{10}}\),\(\displaystyle {2^{12}}\),\(\displaystyle {2^{14}}\)},
y grid style={white!69.0196078431373!black},
ylabel={P(Retrieval Error)},
ymin=-0.0762451361867704, ymax=1.05114785992218,
ytick style={color=black},
ytick={-0.2,0,0.2,0.4,0.6,0.8,1,1.2},
yticklabels={−0.2,0.0,0.2,0.4,0.6,0.8,1.0,1.2}
]
\path [draw=red, fill=red, opacity=0.2]
(axis cs:8,0.643003831361382)
--(axis cs:8,0.381996168638618)
--(axis cs:12,0.457739104119121)
--(axis cs:16,0.711372875703132)
--(axis cs:24,0.797815938765583)
--(axis cs:32,0.884245016941274)
--(axis cs:64,0.9375)
--(axis cs:96,0.973786300635509)
--(axis cs:128,0.969551823467961)
--(axis cs:192,0.97997964119032)
--(axis cs:256,0.987718630269957)
--(axis cs:384,0.9953125)
--(axis cs:512,0.995084931369158)
--(axis cs:1024,0.997112826175171)
--(axis cs:2056,0.998479255877363)
--(axis cs:2056,0.999769771359991)
--(axis cs:2056,0.999769771359991)
--(axis cs:1024,0.999762173824829)
--(axis cs:512,0.998665068630842)
--(axis cs:384,0.998958333333333)
--(axis cs:256,0.995875119730043)
--(axis cs:192,0.999187025476347)
--(axis cs:128,0.999198176532039)
--(axis cs:96,0.992880366031158)
--(axis cs:64,0.984375)
--(axis cs:32,0.959504983058726)
--(axis cs:24,0.927184061234417)
--(axis cs:16,0.851127124296868)
--(axis cs:12,0.808927562547546)
--(axis cs:8,0.643003831361382)
--cycle;

\path [draw=color0, fill=color0, opacity=0.2]
(axis cs:8,0.2)
--(axis cs:8,-0.025)
--(axis cs:12,0.0903708798216374)
--(axis cs:16,0.308422025984384)
--(axis cs:24,0.516666666666667)
--(axis cs:32,0.708242839075821)
--(axis cs:64,0.857340713443485)
--(axis cs:96,0.925)
--(axis cs:128,0.944777001121364)
--(axis cs:192,0.974407105652491)
--(axis cs:256,0.981979396113117)
--(axis cs:384,0.985033279933791)
--(axis cs:512,0.992407355061125)
--(axis cs:1024,0.995835047788997)
--(axis cs:2056,0.998735408560311)
--(axis cs:2056,0.999902723735408)
--(axis cs:2056,0.999902723735408)
--(axis cs:1024,0.998305577211003)
--(axis cs:512,0.997045769938875)
--(axis cs:384,0.996737553399543)
--(axis cs:256,0.993801853886883)
--(axis cs:192,0.993301227680842)
--(axis cs:128,0.981785498878636)
--(axis cs:96,0.9625)
--(axis cs:64,0.945784286556515)
--(axis cs:32,0.804257160924179)
--(axis cs:24,0.716666666666667)
--(axis cs:16,0.504077974015616)
--(axis cs:12,0.359629120178363)
--(axis cs:8,0.2)
--cycle;

\path [draw=blue, fill=blue, opacity=0.2, line width=0pt]
(axis cs:8,0.193541434669349)
--(axis cs:8,0.00645856533065146)
--(axis cs:12,0.16258336669336)
--(axis cs:16,0.363126570361673)
--(axis cs:24,0.548125231997318)
--(axis cs:32,0.664371189166073)
--(axis cs:64,0.862656642590418)
--(axis cs:96,0.934346087106958)
--(axis cs:128,0.950659769983279)
--(axis cs:192,0.969154297699918)
--(axis cs:256,0.979431090692302)
--(axis cs:384,0.990706099598841)
--(axis cs:512,0.991286475053548)
--(axis cs:1024,0.99654284966144)
--(axis cs:2056,0.998346303501945)
--(axis cs:2056,0.999902723735408)
--(axis cs:2056,0.999902723735408)
--(axis cs:1024,0.99974621283856)
--(axis cs:512,0.997776024946452)
--(axis cs:384,0.997314733734492)
--(axis cs:256,0.996350159307698)
--(axis cs:192,0.990220702300082)
--(axis cs:128,0.975902730016721)
--(axis cs:96,0.971903912893042)
--(axis cs:64,0.924843357409582)
--(axis cs:32,0.766878810833927)
--(axis cs:24,0.735208101336015)
--(axis cs:16,0.611873429638327)
--(axis cs:12,0.370749966639973)
--(axis cs:8,0.193541434669349)
--cycle;

\path [draw=green!50!black, fill=green!50!black, opacity=0.2, line width=0pt]
(axis cs:8,0.417705098312484)
--(axis cs:8,0.0822949016875158)
--(axis cs:12,0.287859730876659)
--(axis cs:16,0.5125)
--(axis cs:24,0.684630926914794)
--(axis cs:32,0.754448714203472)
--(axis cs:64,0.922361597593048)
--(axis cs:96,0.956424833785209)
--(axis cs:128,0.9484375)
--(axis cs:192,0.974185072934245)
--(axis cs:256,0.982368934091539)
--(axis cs:384,0.987340300071397)
--(axis cs:512,0.995084931369158)
--(axis cs:1024,0.995957937555697)
--(axis cs:2056,0.998556003037916)
--(axis cs:2056,0.99979030046403)
--(axis cs:2056,0.99979030046403)
--(axis cs:1024,0.999745187444303)
--(axis cs:512,0.998665068630842)
--(axis cs:384,0.995993033261937)
--(axis cs:256,0.991068565908461)
--(axis cs:192,0.988314927065755)
--(axis cs:128,0.9859375)
--(axis cs:96,0.981075166214791)
--(axis cs:64,0.962013402406952)
--(axis cs:32,0.908051285796528)
--(axis cs:24,0.790369073085206)
--(axis cs:16,0.6625)
--(axis cs:12,0.528806935790008)
--(axis cs:8,0.417705098312484)
--cycle;

\addplot [semithick, red]
table {%
8 0.5125
12 0.633333333333333
16 0.78125
24 0.8625
32 0.921875
64 0.9609375
96 0.983333333333333
128 0.984375
192 0.989583333333333
256 0.991796875
384 0.997135416666667
512 0.996875
1024 0.9984375
2056 0.999124513618677
};
\addlegendentry{HRR}
\addplot [semithick, color0, dashed]
table {%
8 0.0875
12 0.225
16 0.40625
24 0.616666666666667
32 0.75625
64 0.9015625
96 0.94375
128 0.96328125
192 0.983854166666667
256 0.987890625
384 0.990885416666667
512 0.9947265625
1024 0.9970703125
2056 0.99931906614786
};
\addlegendentry{HRR w/ Proj}
\addplot [semithick, blue, dash pattern=on 1pt off 3pt on 3pt off 3pt]
table {%
8 0.1
12 0.266666666666667
16 0.4875
24 0.641666666666667
32 0.715625
64 0.89375
96 0.953125
128 0.96328125
192 0.9796875
256 0.987890625
384 0.994010416666667
512 0.99453125
1024 0.99814453125
2056 0.999124513618677
};
\addlegendentry{VTB}
\addplot [semithick, green!50!black, dotted]
table {%
8 0.25
12 0.408333333333333
16 0.5875
24 0.7375
32 0.83125
64 0.9421875
96 0.96875
128 0.9671875
192 0.98125
256 0.98671875
384 0.991666666666667
512 0.996875
1024 0.9978515625
2056 0.999173151750973
};
\addlegendentry{MAP-C}
\end{axis}

\end{tikzpicture}
        }
    }
     \subfigure[$d=121$]{
         \centering
        \adjustbox{max width=0.3\textwidth}{
        \begin{tikzpicture}

\definecolor{color0}{rgb}{1,0.647058823529412,0}

\begin{axis}[
legend cell align={left},
legend style={fill opacity=0.8, draw opacity=1, text opacity=1, at={(0.97,0.03)}, anchor=south east, draw=white!80!black},
log basis x={2},
tick align=outside,
tick pos=left,
title={d=121 Dimensions},
x grid style={white!69.0196078431373!black},
xlabel={Number of Vector Pairs and Distractors},
xmin=6.06168453445102, xmax=2713.43714878584,
xmode=log,
xtick style={color=black},
xtick={1,4,16,64,256,1024,4096,16384},
xticklabels={\(\displaystyle {2^{0}}\),\(\displaystyle {2^{2}}\),\(\displaystyle {2^{4}}\),\(\displaystyle {2^{6}}\),\(\displaystyle {2^{8}}\),\(\displaystyle {2^{10}}\),\(\displaystyle {2^{12}}\),\(\displaystyle {2^{14}}\)},
y grid style={white!69.0196078431373!black},
ylabel={P(Retrieval Error)},
ymin=-0.0762432287486613, ymax=1.05110780372189,
ytick style={color=black},
ytick={-0.2,0,0.2,0.4,0.6,0.8,1,1.2},
yticklabels={−0.2,0.0,0.2,0.4,0.6,0.8,1.0,1.2}
]
\path [draw=red, fill=red, opacity=0.2]
(axis cs:8,0.659629120178363)
--(axis cs:8,0.390370879821637)
--(axis cs:12,0.423704213154971)
--(axis cs:16,0.603527526411483)
--(axis cs:24,0.696448751890197)
--(axis cs:32,0.783942633914947)
--(axis cs:64,0.890385322404306)
--(axis cs:96,0.956645313175194)
--(axis cs:128,0.963384404481162)
--(axis cs:192,0.983420750225938)
--(axis cs:256,0.991907110200359)
--(axis cs:384,0.990918140081501)
--(axis cs:512,0.994044922656876)
--(axis cs:1024,0.996619800026774)
--(axis cs:2056,0.998305817267792)
--(axis cs:2056,0.999748657440379)
--(axis cs:2056,0.999748657440379)
--(axis cs:1024,0.999864574973226)
--(axis cs:512,0.998923827343124)
--(axis cs:384,0.997102693251833)
--(axis cs:256,0.998717889799641)
--(axis cs:192,0.995745916440729)
--(axis cs:128,0.992865595518838)
--(axis cs:96,0.989188020158139)
--(axis cs:64,0.972114677595694)
--(axis cs:32,0.897307366085053)
--(axis cs:24,0.836884581443136)
--(axis cs:16,0.821472473588517)
--(axis cs:12,0.692962453511696)
--(axis cs:8,0.659629120178363)
--cycle;

\path [draw=color0, fill=color0, opacity=0.2]
(axis cs:8,0.05)
--(axis cs:8,-0.025)
--(axis cs:12,-0.0131881307912987)
--(axis cs:16,0.0875)
--(axis cs:24,0.343580586547758)
--(axis cs:32,0.46568752502002)
--(axis cs:64,0.77568589321038)
--(axis cs:96,0.865285670211218)
--(axis cs:128,0.903222847733579)
--(axis cs:192,0.958606039276173)
--(axis cs:256,0.971391445282326)
--(axis cs:384,0.988861893329618)
--(axis cs:512,0.990319402350342)
--(axis cs:1024,0.995975922303151)
--(axis cs:2056,0.99820981281737)
--(axis cs:2056,0.999455556832435)
--(axis cs:2056,0.999455556832435)
--(axis cs:1024,0.998360015196849)
--(axis cs:512,0.995618097649658)
--(axis cs:384,0.994471440003716)
--(axis cs:256,0.989546054717674)
--(axis cs:192,0.986185627390493)
--(axis cs:128,0.953027152266421)
--(axis cs:96,0.926380996455448)
--(axis cs:64,0.84618910678962)
--(axis cs:32,0.62181247497998)
--(axis cs:24,0.523086080118908)
--(axis cs:16,0.2625)
--(axis cs:12,0.0631881307912987)
--(axis cs:8,0.05)
--cycle;

\path [draw=blue, fill=blue, opacity=0.2, line width=0pt]
(axis cs:8,0.094782196186948)
--(axis cs:8,-0.019782196186948)
--(axis cs:12,0)
--(axis cs:16,0.132685439910819)
--(axis cs:24,0.391666666666667)
--(axis cs:32,0.489005360847865)
--(axis cs:64,0.809078718593098)
--(axis cs:96,0.879166666666667)
--(axis cs:128,0.906431095739492)
--(axis cs:192,0.957579369484621)
--(axis cs:256,0.972463319714094)
--(axis cs:384,0.985846523027891)
--(axis cs:512,0.989453125)
--(axis cs:1024,0.996229175026774)
--(axis cs:2056,0.998072932910904)
--(axis cs:2056,0.999592436738901)
--(axis cs:2056,0.999592436738901)
--(axis cs:1024,0.999473949973226)
--(axis cs:512,0.995703125)
--(axis cs:384,0.994882643638775)
--(axis cs:256,0.988474180285906)
--(axis cs:192,0.979920630515379)
--(axis cs:128,0.946693904260508)
--(axis cs:96,0.914583333333333)
--(axis cs:64,0.875296281406902)
--(axis cs:32,0.617244639152135)
--(axis cs:24,0.466666666666667)
--(axis cs:16,0.267314560089181)
--(axis cs:12,0.25)
--(axis cs:8,0.094782196186948)
--cycle;

\path [draw=green!50!black, fill=green!50!black, opacity=0.2, line width=0pt]
(axis cs:8,0.175)
--(axis cs:8,0.025)
--(axis cs:12,0.170972376887101)
--(axis cs:16,0.313897569783555)
--(axis cs:24,0.533712907082472)
--(axis cs:32,0.666902311618831)
--(axis cs:64,0.8439737866338)
--(axis cs:96,0.905229175006673)
--(axis cs:128,0.944181554226932)
--(axis cs:192,0.971555600142794)
--(axis cs:256,0.980600818832152)
--(axis cs:384,0.988020833333333)
--(axis cs:512,0.991870996220349)
--(axis cs:1024,0.996416314603838)
--(axis cs:2056,0.998595605929619)
--(axis cs:2056,0.999750697572327)
--(axis cs:2056,0.999750697572327)
--(axis cs:1024,0.999286810396162)
--(axis cs:512,0.997972753779651)
--(axis cs:384,0.992708333333333)
--(axis cs:256,0.989711681167848)
--(axis cs:192,0.988861066523873)
--(axis cs:128,0.977693445773068)
--(axis cs:96,0.957270824993327)
--(axis cs:64,0.9310262133662)
--(axis cs:32,0.795597688381169)
--(axis cs:24,0.716287092917528)
--(axis cs:16,0.536102430216445)
--(axis cs:12,0.295694289779566)
--(axis cs:8,0.175)
--cycle;

\addplot [semithick, red]
table {%
8 0.525
12 0.558333333333333
16 0.7125
24 0.766666666666667
32 0.840625
64 0.93125
96 0.972916666666667
128 0.978125
192 0.989583333333333
256 0.9953125
384 0.994010416666667
512 0.996484375
1024 0.9982421875
2056 0.999027237354085
};
\addlegendentry{HRR}
\addplot [semithick, color0, dashed]
table {%
8 0.0125
12 0.025
16 0.175
24 0.433333333333333
32 0.54375
64 0.8109375
96 0.895833333333333
128 0.928125
192 0.972395833333333
256 0.98046875
384 0.991666666666667
512 0.99296875
1024 0.99716796875
2056 0.998832684824903
};
\addlegendentry{HRR w/ Proj}
\addplot [semithick, blue, dash pattern=on 1pt off 3pt on 3pt off 3pt]
table {%
8 0.0375
12 0.125
16 0.2
24 0.429166666666667
32 0.553125
64 0.8421875
96 0.896875
128 0.9265625
192 0.96875
256 0.98046875
384 0.990364583333333
512 0.992578125
1024 0.9978515625
2056 0.998832684824903
};
\addlegendentry{VTB}
\addplot [semithick, green!50!black, dotted]
table {%
8 0.1
12 0.233333333333333
16 0.425
24 0.625
32 0.73125
64 0.8875
96 0.93125
128 0.9609375
192 0.980208333333333
256 0.98515625
384 0.990364583333333
512 0.994921875
1024 0.9978515625
2056 0.999173151750973
};
\addlegendentry{MAP-C}
\end{axis}

\end{tikzpicture}
        }
    }
     \subfigure[$d=256$]{
         \centering
        \adjustbox{max width=0.3\textwidth}{
        \begin{tikzpicture}

\definecolor{color0}{rgb}{1,0.647058823529412,0}

\begin{axis}[
legend cell align={left},
legend style={fill opacity=0.8, draw opacity=1, text opacity=1, at={(0.97,0.03)}, anchor=south east, draw=white!80!black},
log basis x={2},
tick align=outside,
tick pos=left,
title={d=256 Dimensions},
x grid style={white!69.0196078431373!black},
xlabel={Number of Vector Pairs and Distractors},
xmin=6.06168453445102, xmax=2713.43714878584,
xmode=log,
xtick style={color=black},
xtick={1,4,16,64,256,1024,4096,16384},
xticklabels={\(\displaystyle {2^{0}}\),\(\displaystyle {2^{2}}\),\(\displaystyle {2^{4}}\),\(\displaystyle {2^{6}}\),\(\displaystyle {2^{8}}\),\(\displaystyle {2^{10}}\),\(\displaystyle {2^{12}}\),\(\displaystyle {2^{14}}\)},
y grid style={white!69.0196078431373!black},
ylabel={P(Retrieval Error)},
ymin=-0.0946621972046623, ymax=1.05204697600388,
ytick style={color=black},
ytick={-0.2,0,0.2,0.4,0.6,0.8,1,1.2},
yticklabels={−0.2,0.0,0.2,0.4,0.6,0.8,1.0,1.2}
]
\path [draw=red, fill=red, opacity=0.2]
(axis cs:8,0.35)
--(axis cs:8,0.05)
--(axis cs:12,0.234317141532216)
--(axis cs:16,0.3625)
--(axis cs:24,0.607986711843397)
--(axis cs:32,0.748881822838748)
--(axis cs:64,0.8875)
--(axis cs:96,0.940483824244104)
--(axis cs:128,0.933538041011168)
--(axis cs:192,0.980208333333333)
--(axis cs:256,0.988033886882535)
--(axis cs:384,0.991710375112969)
--(axis cs:512,0.9953125)
--(axis cs:1024,0.9966796875)
--(axis cs:2056,0.999006129322371)
--(axis cs:2056,0.999923831767123)
--(axis cs:2056,0.999923831767123)
--(axis cs:1024,0.998828125)
--(axis cs:512,0.999609375)
--(axis cs:384,0.997872958220364)
--(axis cs:256,0.993216113117465)
--(axis cs:192,0.992708333333333)
--(axis cs:128,0.966461958988832)
--(axis cs:96,0.963682842422563)
--(axis cs:64,0.946875)
--(axis cs:32,0.819868177161252)
--(axis cs:24,0.808679954823269)
--(axis cs:16,0.65)
--(axis cs:12,0.515682858467784)
--(axis cs:8,0.35)
--cycle;

\path [draw=color0, fill=color0, opacity=0.2]
(axis cs:8,0)
--(axis cs:8,0)
--(axis cs:12,0)
--(axis cs:16,-0.0039578098794425)
--(axis cs:24,0.0618118692087013)
--(axis cs:32,0.196320478503882)
--(axis cs:64,0.55366703505658)
--(axis cs:96,0.72515171603374)
--(axis cs:128,0.830070892030423)
--(axis cs:192,0.906550089085356)
--(axis cs:256,0.946414021472558)
--(axis cs:384,0.970662393794868)
--(axis cs:512,0.984956057735449)
--(axis cs:1024,0.993831528184579)
--(axis cs:2056,0.997630200246921)
--(axis cs:2056,0.999548788079928)
--(axis cs:2056,0.999548788079928)
--(axis cs:1024,0.995621596815421)
--(axis cs:512,0.992778317264551)
--(axis cs:384,0.981941772871799)
--(axis cs:256,0.974679728527442)
--(axis cs:192,0.942408244247978)
--(axis cs:128,0.893366607969577)
--(axis cs:96,0.839431617299594)
--(axis cs:64,0.64320796494342)
--(axis cs:32,0.284929521496118)
--(axis cs:24,0.138188130791299)
--(axis cs:16,0.0789578098794425)
--(axis cs:12,0)
--(axis cs:8,0)
--cycle;

\path [draw=blue, fill=blue, opacity=0.2, line width=0pt]
(axis cs:8,0)
--(axis cs:8,0)
--(axis cs:12,-0.0166666666666667)
--(axis cs:16,-0.0164578098794425)
--(axis cs:24,0.0706255182453296)
--(axis cs:32,0.174152945464625)
--(axis cs:64,0.571101722769013)
--(axis cs:96,0.76451396587839)
--(axis cs:128,0.852855285109486)
--(axis cs:192,0.929757794628774)
--(axis cs:256,0.952071508415352)
--(axis cs:384,0.970833333333333)
--(axis cs:512,0.980123398934873)
--(axis cs:1024,0.993435640583414)
--(axis cs:2056,0.997762645914397)
--(axis cs:2056,0.999124513618677)
--(axis cs:2056,0.999124513618677)
--(axis cs:1024,0.998361234416586)
--(axis cs:512,0.990189101065127)
--(axis cs:384,0.985416666666667)
--(axis cs:256,0.963553491584648)
--(axis cs:192,0.95670053870456)
--(axis cs:128,0.889332214890514)
--(axis cs:96,0.827152700788277)
--(axis cs:64,0.647648277230987)
--(axis cs:32,0.313347054535375)
--(axis cs:24,0.15437448175467)
--(axis cs:16,0.0664578098794425)
--(axis cs:12,0.0333333333333333)
--(axis cs:8,0)
--cycle;

\path [draw=green!50!black, fill=green!50!black, opacity=0.2, line width=0pt]
(axis cs:8,0.117539052967911)
--(axis cs:8,-0.0425390529679106)
--(axis cs:12,-0.0219437465059233)
--(axis cs:16,0.104959027401664)
--(axis cs:24,0.2730378726934)
--(axis cs:32,0.451262756430421)
--(axis cs:64,0.72554203505658)
--(axis cs:96,0.867070315677363)
--(axis cs:128,0.907116370090219)
--(axis cs:192,0.956953588207982)
--(axis cs:256,0.965987770505584)
--(axis cs:384,0.979752946829668)
--(axis cs:512,0.989948010681129)
--(axis cs:1024,0.995643237526774)
--(axis cs:2056,0.998823052576499)
--(axis cs:2056,0.999815079719221)
--(axis cs:2056,0.999815079719221)
--(axis cs:1024,0.998888012473226)
--(axis cs:512,0.997161364318871)
--(axis cs:384,0.992122053170332)
--(axis cs:256,0.982449729494416)
--(axis cs:192,0.970129745125351)
--(axis cs:128,0.942883629909781)
--(axis cs:96,0.920429684322637)
--(axis cs:64,0.81508296494342)
--(axis cs:32,0.573737243569579)
--(axis cs:24,0.4519621273066)
--(axis cs:16,0.257540972598336)
--(axis cs:12,0.08861041317259)
--(axis cs:8,0.117539052967911)
--cycle;

\addplot [semithick, red]
table {%
8 0.2
12 0.375
16 0.50625
24 0.708333333333333
32 0.784375
64 0.9171875
96 0.952083333333333
128 0.95
192 0.986458333333333
256 0.990625
384 0.994791666666667
512 0.9974609375
1024 0.99775390625
2056 0.999464980544747
};
\addlegendentry{HRR}
\addplot [semithick, color0, dashed]
table {%
8 0
12 0
16 0.0375
24 0.1
32 0.240625
64 0.5984375
96 0.782291666666667
128 0.86171875
192 0.924479166666667
256 0.960546875
384 0.976302083333333
512 0.9888671875
1024 0.9947265625
2056 0.998589494163424
};
\addlegendentry{HRR w/ Proj}
\addplot [semithick, blue, dash pattern=on 1pt off 3pt on 3pt off 3pt]
table {%
8 0
12 0.00833333333333333
16 0.025
24 0.1125
32 0.24375
64 0.609375
96 0.795833333333333
128 0.87109375
192 0.943229166666667
256 0.9578125
384 0.978125
512 0.98515625
1024 0.9958984375
2056 0.998443579766537
};
\addlegendentry{VTB}
\addplot [semithick, green!50!black, dotted]
table {%
8 0.0375
12 0.0333333333333333
16 0.18125
24 0.3625
32 0.5125
64 0.7703125
96 0.89375
128 0.925
192 0.963541666666667
256 0.97421875
384 0.9859375
512 0.9935546875
1024 0.997265625
2056 0.99931906614786
};
\addlegendentry{MAP-C}
\end{axis}

\end{tikzpicture}
        }
    }
     \subfigure[$d=484$]{
         \centering
        \adjustbox{max width=0.3\textwidth}{
        \begin{tikzpicture}

\definecolor{color0}{rgb}{1,0.647058823529412,0}

\begin{axis}[
legend cell align={left},
legend style={fill opacity=0.8, draw opacity=1, text opacity=1, at={(0.97,0.03)}, anchor=south east, draw=white!80!black},
log basis x={2},
tick align=outside,
tick pos=left,
title={d=484 Dimensions},
x grid style={white!69.0196078431373!black},
xlabel={Number of Vector Pairs and Distractors},
xmin=6.06168453445102, xmax=2713.43714878584,
xmode=log,
xtick style={color=black},
xtick={1,4,16,64,256,1024,4096,16384},
xticklabels={\(\displaystyle {2^{0}}\),\(\displaystyle {2^{2}}\),\(\displaystyle {2^{4}}\),\(\displaystyle {2^{6}}\),\(\displaystyle {2^{8}}\),\(\displaystyle {2^{10}}\),\(\displaystyle {2^{12}}\),\(\displaystyle {2^{14}}\)},
y grid style={white!69.0196078431373!black},
ylabel={P(Retrieval Error)},
ymin=-0.058737226962755, ymax=1.05014843288452,
ytick style={color=black},
ytick={-0.2,0,0.2,0.4,0.6,0.8,1,1.2},
yticklabels={−0.2,0.0,0.2,0.4,0.6,0.8,1.0,1.2}
]
\path [draw=red, fill=red, opacity=0.2]
(axis cs:8,0.255424764150708)
--(axis cs:8,0.0195752358492925)
--(axis cs:12,0.104014295594307)
--(axis cs:16,0.248814874372393)
--(axis cs:24,0.429097241501636)
--(axis cs:32,0.65)
--(axis cs:64,0.836068011145048)
--(axis cs:96,0.914583333333333)
--(axis cs:128,0.939125100321286)
--(axis cs:192,0.972314846655519)
--(axis cs:256,0.977002210122251)
--(axis cs:384,0.984375)
--(axis cs:512,0.992738779906493)
--(axis cs:1024,0.996405639797376)
--(axis cs:2056,0.998407211717662)
--(axis cs:2056,0.9997445392551)
--(axis cs:2056,0.9997445392551)
--(axis cs:1024,0.998906860202624)
--(axis cs:512,0.998276845093507)
--(axis cs:384,0.997395833333333)
--(axis cs:256,0.986279039877749)
--(axis cs:192,0.989143486677814)
--(axis cs:128,0.978062399678714)
--(axis cs:96,0.952083333333333)
--(axis cs:64,0.892056988854951)
--(axis cs:32,0.79375)
--(axis cs:24,0.679236091831697)
--(axis cs:16,0.513685125627607)
--(axis cs:12,0.445985704405693)
--(axis cs:8,0.255424764150708)
--cycle;

\path [draw=color0, fill=color0, opacity=0.2]
(axis cs:8,0)
--(axis cs:8,0)
--(axis cs:12,0)
--(axis cs:16,0)
--(axis cs:24,-0.00833333333333333)
--(axis cs:32,0.0162179697627917)
--(axis cs:64,0.308119517965652)
--(axis cs:96,0.518648494608261)
--(axis cs:128,0.68884041308792)
--(axis cs:192,0.838496419537911)
--(axis cs:256,0.907059187465805)
--(axis cs:384,0.952564220400718)
--(axis cs:512,0.970195196243589)
--(axis cs:1024,0.990774325141252)
--(axis cs:2056,0.996052934153425)
--(axis cs:2056,0.998305042500271)
--(axis cs:2056,0.998305042500271)
--(axis cs:1024,0.994967862358748)
--(axis cs:512,0.983320428756411)
--(axis cs:384,0.966185779599282)
--(axis cs:256,0.928878312534195)
--(axis cs:192,0.886503580462089)
--(axis cs:128,0.73303458691208)
--(axis cs:96,0.604268172058405)
--(axis cs:64,0.379380482034348)
--(axis cs:32,0.0650320302372083)
--(axis cs:24,0.0166666666666667)
--(axis cs:16,0)
--(axis cs:12,0)
--(axis cs:8,0)
--cycle;

\path [draw=blue, fill=blue, opacity=0.2, line width=0pt]
(axis cs:8,0)
--(axis cs:8,0)
--(axis cs:12,0)
--(axis cs:16,0)
--(axis cs:24,-0.00711751638541404)
--(axis cs:32,0.0102568816028708)
--(axis cs:64,0.322855434502778)
--(axis cs:96,0.546018885768719)
--(axis cs:128,0.697368145091628)
--(axis cs:192,0.830182941925236)
--(axis cs:256,0.893591219801507)
--(axis cs:384,0.948532724545791)
--(axis cs:512,0.967456440183376)
--(axis cs:1024,0.990460786944114)
--(axis cs:2056,0.996421489512597)
--(axis cs:2056,0.998228315934874)
--(axis cs:2056,0.998228315934874)
--(axis cs:1024,0.994109525555886)
--(axis cs:512,0.981371684816624)
--(axis cs:384,0.965529775454209)
--(axis cs:256,0.932190030198493)
--(axis cs:192,0.872942058074764)
--(axis cs:128,0.746381854908372)
--(axis cs:96,0.639397780897948)
--(axis cs:64,0.411519565497222)
--(axis cs:32,0.0647431183971292)
--(axis cs:24,0.0487841830520807)
--(axis cs:16,0)
--(axis cs:12,0)
--(axis cs:8,0)
--cycle;

\path [draw=green!50!black, fill=green!50!black, opacity=0.2, line width=0pt]
(axis cs:8,0)
--(axis cs:8,0)
--(axis cs:12,0)
--(axis cs:16,-0.00561862178478973)
--(axis cs:24,0.0404880964288096)
--(axis cs:32,0.167600423853725)
--(axis cs:64,0.589207424852895)
--(axis cs:96,0.758574794319323)
--(axis cs:128,0.858127111582893)
--(axis cs:192,0.923024812449075)
--(axis cs:256,0.945032955709687)
--(axis cs:384,0.976308287249114)
--(axis cs:512,0.980116046171582)
--(axis cs:1024,0.993406417526496)
--(axis cs:2056,0.997746580604094)
--(axis cs:2056,0.999529683987345)
--(axis cs:2056,0.999529683987345)
--(axis cs:1024,0.997413894973504)
--(axis cs:512,0.992149578828418)
--(axis cs:384,0.987233379417553)
--(axis cs:256,0.962779544290313)
--(axis cs:192,0.953016854217592)
--(axis cs:128,0.904372888417107)
--(axis cs:96,0.812258539014011)
--(axis cs:64,0.673292575147105)
--(axis cs:32,0.338649576146275)
--(axis cs:24,0.15951190357119)
--(axis cs:16,0.0556186217847897)
--(axis cs:12,0)
--(axis cs:8,0)
--cycle;

\addplot [semithick, red]
table {%
8 0.1375
12 0.275
16 0.38125
24 0.554166666666667
32 0.721875
64 0.8640625
96 0.933333333333333
128 0.95859375
192 0.980729166666667
256 0.981640625
384 0.990885416666667
512 0.9955078125
1024 0.99765625
2056 0.999075875486381
};
\addlegendentry{HRR}
\addplot [semithick, color0, dashed]
table {%
8 0
12 0
16 0
24 0.00416666666666667
32 0.040625
64 0.34375
96 0.561458333333333
128 0.7109375
192 0.8625
256 0.91796875
384 0.959375
512 0.9767578125
1024 0.99287109375
2056 0.997178988326848
};
\addlegendentry{HRR w/ Proj}
\addplot [semithick, blue, dash pattern=on 1pt off 3pt on 3pt off 3pt]
table {%
8 0
12 0
16 0
24 0.0208333333333333
32 0.0375
64 0.3671875
96 0.592708333333333
128 0.721875
192 0.8515625
256 0.912890625
384 0.95703125
512 0.9744140625
1024 0.99228515625
2056 0.997324902723735
};
\addlegendentry{VTB}
\addplot [semithick, green!50!black, dotted]
table {%
8 0
12 0
16 0.025
24 0.1
32 0.253125
64 0.63125
96 0.785416666666667
128 0.88125
192 0.938020833333333
256 0.95390625
384 0.981770833333333
512 0.9861328125
1024 0.99541015625
2056 0.99863813229572
};
\addlegendentry{MAP-C}
\end{axis}

\end{tikzpicture}
        }
    }
     \subfigure[$d=1024$]{
         \centering
        \adjustbox{max width=0.3\textwidth}{
        \begin{tikzpicture}

\definecolor{color0}{rgb}{1,0.647058823529412,0}

\begin{axis}[
legend cell align={left},
legend style={fill opacity=0.8, draw opacity=1, text opacity=1, at={(0.03,0.97)}, anchor=north west, draw=white!80!black},
log basis x={2},
tick align=outside,
tick pos=left,
title={d=1024 Dimensions},
x grid style={white!69.0196078431373!black},
xlabel={Number of Vector Pairs and Distractors},
xmin=6.06168453445102, xmax=2713.43714878584,
xmode=log,
xtick style={color=black},
xtick={1,4,16,64,256,1024,4096,16384},
xticklabels={\(\displaystyle {2^{0}}\),\(\displaystyle {2^{2}}\),\(\displaystyle {2^{4}}\),\(\displaystyle {2^{6}}\),\(\displaystyle {2^{8}}\),\(\displaystyle {2^{10}}\),\(\displaystyle {2^{12}}\),\(\displaystyle {2^{14}}\)},
y grid style={white!69.0196078431373!black},
ylabel={P(Retrieval Error)},
ymin=-0.0762334789776647, ymax=1.05090305853096,
ytick style={color=black},
ytick={-0.2,0,0.2,0.4,0.6,0.8,1,1.2},
yticklabels={−0.2,0.0,0.2,0.4,0.6,0.8,1.0,1.2}
]
\path [draw=red, fill=red, opacity=0.2]
(axis cs:8,0.05)
--(axis cs:8,-0.025)
--(axis cs:12,0.0376390435537676)
--(axis cs:16,0.116027526411483)
--(axis cs:24,0.328306018267776)
--(axis cs:32,0.425)
--(axis cs:64,0.739344769856279)
--(axis cs:96,0.83125)
--(axis cs:128,0.897171484881396)
--(axis cs:192,0.938374521079827)
--(axis cs:256,0.970081538952354)
--(axis cs:384,0.978380767850978)
--(axis cs:512,0.988671875)
--(axis cs:1024,0.995495652783535)
--(axis cs:2056,0.998482171419468)
--(axis cs:2056,0.999669579553294)
--(axis cs:2056,0.999669579553294)
--(axis cs:1024,0.998644972216465)
--(axis cs:512,0.993359375)
--(axis cs:384,0.992973398815688)
--(axis cs:256,0.986168461047646)
--(axis cs:192,0.971000478920173)
--(axis cs:128,0.921578515118604)
--(axis cs:96,0.889583333333333)
--(axis cs:64,0.816905230143721)
--(axis cs:32,0.56875)
--(axis cs:24,0.430027315065557)
--(axis cs:16,0.333972473588517)
--(axis cs:12,0.162360956446232)
--(axis cs:8,0.05)
--cycle;

\path [draw=color0, fill=color0, opacity=0.2]
(axis cs:8,0)
--(axis cs:8,0)
--(axis cs:12,0)
--(axis cs:16,0)
--(axis cs:24,0)
--(axis cs:32,-0.00625)
--(axis cs:64,0.0337561280400451)
--(axis cs:96,0.215759175470494)
--(axis cs:128,0.381312600321286)
--(axis cs:192,0.608109712910509)
--(axis cs:256,0.756481120060737)
--(axis cs:384,0.879306849350144)
--(axis cs:512,0.930631570476065)
--(axis cs:1024,0.981660902334202)
--(axis cs:2056,0.995088707346138)
--(axis cs:2056,0.997129191486547)
--(axis cs:2056,0.997129191486547)
--(axis cs:1024,0.987284410165798)
--(axis cs:512,0.948274679523935)
--(axis cs:384,0.908193150649857)
--(axis cs:256,0.795862629939263)
--(axis cs:192,0.676265287089491)
--(axis cs:128,0.420249899678714)
--(axis cs:96,0.288407491196172)
--(axis cs:64,0.0974938719599549)
--(axis cs:32,0.0125)
--(axis cs:24,0)
--(axis cs:16,0)
--(axis cs:12,0)
--(axis cs:8,0)
--cycle;

\path [draw=blue, fill=blue, opacity=0.2, line width=0pt]
(axis cs:8,0)
--(axis cs:8,0)
--(axis cs:12,0)
--(axis cs:16,0)
--(axis cs:24,0)
--(axis cs:32,0)
--(axis cs:64,0.0445840854490322)
--(axis cs:96,0.20116856476128)
--(axis cs:128,0.346854444347662)
--(axis cs:192,0.634618572585274)
--(axis cs:256,0.760840012705956)
--(axis cs:384,0.888048283686721)
--(axis cs:512,0.932655773645759)
--(axis cs:1024,0.980172027090636)
--(axis cs:2056,0.994036936029532)
--(axis cs:2056,0.99730547642183)
--(axis cs:2056,0.99730547642183)
--(axis cs:1024,0.987796722909364)
--(axis cs:512,0.945469226354241)
--(axis cs:384,0.907785049646613)
--(axis cs:256,0.817284987294044)
--(axis cs:192,0.678923094081393)
--(axis cs:128,0.465645555652338)
--(axis cs:96,0.277998101905387)
--(axis cs:64,0.102290914550968)
--(axis cs:32,0)
--(axis cs:24,0)
--(axis cs:16,0)
--(axis cs:12,0)
--(axis cs:8,0)
--cycle;

\path [draw=green!50!black, fill=green!50!black, opacity=0.2, line width=0pt]
(axis cs:8,0)
--(axis cs:8,0)
--(axis cs:12,0)
--(axis cs:16,0)
--(axis cs:24,-0.00659406539564933)
--(axis cs:32,0.0162179697627917)
--(axis cs:64,0.232834825985825)
--(axis cs:96,0.519541842139129)
--(axis cs:128,0.661094142590418)
--(axis cs:192,0.828329228641409)
--(axis cs:256,0.890821548704585)
--(axis cs:384,0.94449946440346)
--(axis cs:512,0.9703125)
--(axis cs:1024,0.989239294912069)
--(axis cs:2056,0.99727626459144)
--(axis cs:2056,0.998443579766537)
--(axis cs:2056,0.998443579766537)
--(axis cs:1024,0.993377892587931)
--(axis cs:512,0.9796875)
--(axis cs:384,0.96800053559654)
--(axis cs:256,0.921678451295415)
--(axis cs:192,0.870629104691924)
--(axis cs:128,0.723280857409582)
--(axis cs:96,0.628374824527537)
--(axis cs:64,0.342165174014175)
--(axis cs:32,0.0650320302372083)
--(axis cs:24,0.0315940653956493)
--(axis cs:16,0)
--(axis cs:12,0)
--(axis cs:8,0)
--cycle;

\addplot [semithick, red]
table {%
8 0.0125
12 0.1
16 0.225
24 0.379166666666667
32 0.496875
64 0.778125
96 0.860416666666667
128 0.909375
192 0.9546875
256 0.978125
384 0.985677083333333
512 0.991015625
1024 0.9970703125
2056 0.999075875486381
};
\addlegendentry{HRR}
\addplot [semithick, color0, dashed]
table {%
8 0
12 0
16 0
24 0
32 0.003125
64 0.065625
96 0.252083333333333
128 0.40078125
192 0.6421875
256 0.776171875
384 0.89375
512 0.939453125
1024 0.98447265625
2056 0.996108949416342
};
\addlegendentry{HRR w/ Proj}
\addplot [semithick, blue, dash pattern=on 1pt off 3pt on 3pt off 3pt]
table {%
8 0
12 0
16 0
24 0
32 0
64 0.0734375
96 0.239583333333333
128 0.40625
192 0.656770833333333
256 0.7890625
384 0.897916666666667
512 0.9390625
1024 0.983984375
2056 0.995671206225681
};
\addlegendentry{VTB}
\addplot [semithick, green!50!black, dotted]
table {%
8 0
12 0
16 0
24 0.0125
32 0.040625
64 0.2875
96 0.573958333333333
128 0.6921875
192 0.849479166666667
256 0.90625
384 0.95625
512 0.975
1024 0.99130859375
2056 0.997859922178988
};
\addlegendentry{MAP-C}
\end{axis}

\end{tikzpicture}
        }
    }
     \subfigure[$d=2025$]{
         \centering
        \adjustbox{max width=0.3\textwidth}{
        \begin{tikzpicture}

\definecolor{color0}{rgb}{1,0.647058823529412,0}

\begin{axis}[
legend cell align={left},
legend style={fill opacity=0.8, draw opacity=1, text opacity=1, at={(0.03,0.97)}, anchor=north west, draw=white!80!black},
log basis x={2},
tick align=outside,
tick pos=left,
title={d=2025 Dimensions},
x grid style={white!69.0196078431373!black},
xlabel={Number of Vector Pairs and Distractors},
xmin=6.06168453445102, xmax=2713.43714878584,
xmode=log,
xtick style={color=black},
xtick={1,4,16,64,256,1024,4096,16384},
xticklabels={\(\displaystyle {2^{0}}\),\(\displaystyle {2^{2}}\),\(\displaystyle {2^{4}}\),\(\displaystyle {2^{6}}\),\(\displaystyle {2^{8}}\),\(\displaystyle {2^{10}}\),\(\displaystyle {2^{12}}\),\(\displaystyle {2^{14}}\)},
y grid style={white!69.0196078431373!black},
ylabel={P(Retrieval Error)},
ymin=-0.0762215050020844, ymax=1.05065160504377,
ytick style={color=black},
ytick={-0.2,0,0.2,0.4,0.6,0.8,1,1.2},
yticklabels={−0.2,0.0,0.2,0.4,0.6,0.8,1.0,1.2}
]
\path [draw=red, fill=red, opacity=0.2]
(axis cs:8,0.05)
--(axis cs:8,-0.025)
--(axis cs:12,-0.0131881307912987)
--(axis cs:16,0.00373047351604469)
--(axis cs:24,0.1)
--(axis cs:32,0.216037617924646)
--(axis cs:64,0.580870029289683)
--(axis cs:96,0.745787059876853)
--(axis cs:128,0.819253241627334)
--(axis cs:192,0.902595488521038)
--(axis cs:256,0.94453125)
--(axis cs:384,0.973170576980599)
--(axis cs:512,0.979163741375135)
--(axis cs:1024,0.994076522673075)
--(axis cs:2056,0.998040717078936)
--(axis cs:2056,0.999430100041687)
--(axis cs:2056,0.999430100041687)
--(axis cs:1024,0.998306289826925)
--(axis cs:512,0.990758133624866)
--(axis cs:384,0.983600256352735)
--(axis cs:256,0.9671875)
--(axis cs:192,0.933862844812295)
--(axis cs:128,0.896371758372666)
--(axis cs:96,0.839629606789813)
--(axis cs:64,0.709754970710316)
--(axis cs:32,0.333962382075354)
--(axis cs:24,0.283333333333333)
--(axis cs:16,0.0837695264839553)
--(axis cs:12,0.0631881307912987)
--(axis cs:8,0.05)
--cycle;

\path [draw=color0, fill=color0, opacity=0.2]
(axis cs:8,0)
--(axis cs:8,0)
--(axis cs:12,0)
--(axis cs:16,0)
--(axis cs:24,0)
--(axis cs:32,0)
--(axis cs:64,-0.003125)
--(axis cs:96,0.0097370043307307)
--(axis cs:128,0.085402001121364)
--(axis cs:192,0.285827803009464)
--(axis cs:256,0.476730426357928)
--(axis cs:384,0.72384157021171)
--(axis cs:512,0.82654514745382)
--(axis cs:1024,0.957872239308457)
--(axis cs:2056,0.989396887159533)
--(axis cs:2056,0.992023346303502)
--(axis cs:2056,0.992023346303502)
--(axis cs:1024,0.964198073191543)
--(axis cs:512,0.86173610254618)
--(axis cs:384,0.74022092978829)
--(axis cs:256,0.531082073642072)
--(axis cs:192,0.332922196990536)
--(axis cs:128,0.122410498878636)
--(axis cs:96,0.0423463290026026)
--(axis cs:64,0.009375)
--(axis cs:32,0)
--(axis cs:24,0)
--(axis cs:16,0)
--(axis cs:12,0)
--(axis cs:8,0)
--cycle;

\path [draw=blue, fill=blue, opacity=0.2, line width=0pt]
(axis cs:8,0)
--(axis cs:8,0)
--(axis cs:12,0)
--(axis cs:16,0)
--(axis cs:24,0)
--(axis cs:32,0)
--(axis cs:64,0)
--(axis cs:96,0.0157995864852174)
--(axis cs:128,0.0811757631955242)
--(axis cs:192,0.280705064339851)
--(axis cs:256,0.487188956908122)
--(axis cs:384,0.71197063070868)
--(axis cs:512,0.820293912848166)
--(axis cs:1024,0.954453266021694)
--(axis cs:2056,0.988771290717121)
--(axis cs:2056,0.992648942745914)
--(axis cs:2056,0.992648942745914)
--(axis cs:1024,0.965273296478306)
--(axis cs:512,0.844940462151834)
--(axis cs:384,0.743758535957987)
--(axis cs:256,0.554217293091878)
--(axis cs:192,0.337003268993483)
--(axis cs:128,0.114136736804476)
--(axis cs:96,0.0467004135147826)
--(axis cs:64,0)
--(axis cs:32,0)
--(axis cs:24,0)
--(axis cs:16,0)
--(axis cs:12,0)
--(axis cs:8,0)
--cycle;

\path [draw=green!50!black, fill=green!50!black, opacity=0.2, line width=0pt]
(axis cs:8,0)
--(axis cs:8,0)
--(axis cs:12,0)
--(axis cs:16,0)
--(axis cs:24,0)
--(axis cs:32,0)
--(axis cs:64,0.0570986995393735)
--(axis cs:96,0.202693945154697)
--(axis cs:128,0.362322899394537)
--(axis cs:192,0.62379431274811)
--(axis cs:256,0.752368964738747)
--(axis cs:384,0.887066174096699)
--(axis cs:512,0.929542363754897)
--(axis cs:1024,0.983151936064818)
--(axis cs:2056,0.993852268078447)
--(axis cs:2056,0.996809210520775)
--(axis cs:2056,0.996809210520775)
--(axis cs:1024,0.986184001435182)
--(axis cs:512,0.945457636245103)
--(axis cs:384,0.924392159236634)
--(axis cs:256,0.800756035261253)
--(axis cs:192,0.683497353918556)
--(axis cs:128,0.431427100605462)
--(axis cs:96,0.266056054845303)
--(axis cs:64,0.102276300460626)
--(axis cs:32,0)
--(axis cs:24,0)
--(axis cs:16,0)
--(axis cs:12,0)
--(axis cs:8,0)
--cycle;

\addplot [semithick, red]
table {%
8 0.0125
12 0.025
16 0.04375
24 0.191666666666667
32 0.275
64 0.6453125
96 0.792708333333333
128 0.8578125
192 0.918229166666667
256 0.955859375
384 0.978385416666667
512 0.9849609375
1024 0.99619140625
2056 0.998735408560311
};
\addlegendentry{HRR}
\addplot [semithick, color0, dashed]
table {%
8 0
12 0
16 0
24 0
32 0
64 0.003125
96 0.0260416666666667
128 0.10390625
192 0.309375
256 0.50390625
384 0.73203125
512 0.844140625
1024 0.96103515625
2056 0.990710116731518
};
\addlegendentry{HRR w/ Proj}
\addplot [semithick, blue, dash pattern=on 1pt off 3pt on 3pt off 3pt]
table {%
8 0
12 0
16 0
24 0
32 0
64 0
96 0.03125
128 0.09765625
192 0.308854166666667
256 0.520703125
384 0.727864583333333
512 0.8326171875
1024 0.95986328125
2056 0.990710116731517
};
\addlegendentry{VTB}
\addplot [semithick, green!50!black, dotted]
table {%
8 0
12 0
16 0
24 0
32 0
64 0.0796875
96 0.234375
128 0.396875
192 0.653645833333333
256 0.7765625
384 0.905729166666667
512 0.9375
1024 0.98466796875
2056 0.995330739299611
};
\addlegendentry{MAP-C}
\end{axis}

\end{tikzpicture}
        }
    }
     \subfigure[$d=4096$]{
         \centering
        \adjustbox{max width=0.3\textwidth}{
        \begin{tikzpicture}

\definecolor{color0}{rgb}{1,0.647058823529412,0}

\begin{axis}[
legend cell align={left},
legend style={fill opacity=0.8, draw opacity=1, text opacity=1, at={(0.03,0.97)}, anchor=north west, draw=white!80!black},
log basis x={2},
tick align=outside,
tick pos=left,
title={d=4096 Dimensions},
x grid style={white!69.0196078431373!black},
xlabel={Number of Vector Pairs and Distractors},
xmin=6.06168453445102, xmax=2713.43714878584,
xmode=log,
xtick style={color=black},
xtick={1,4,16,64,256,1024,4096,16384},
xticklabels={\(\displaystyle {2^{0}}\),\(\displaystyle {2^{2}}\),\(\displaystyle {2^{4}}\),\(\displaystyle {2^{6}}\),\(\displaystyle {2^{8}}\),\(\displaystyle {2^{10}}\),\(\displaystyle {2^{12}}\),\(\displaystyle {2^{14}}\)},
y grid style={white!69.0196078431373!black},
ylabel={P(Retrieval Error)},
ymin=-0.0674019492745435, ymax=1.04877426809875,
ytick style={color=black},
ytick={-0.2,0,0.2,0.4,0.6,0.8,1,1.2},
yticklabels={−0.2,0.0,0.2,0.4,0.6,0.8,1.0,1.2}
]
\path [draw=red, fill=red, opacity=0.2]
(axis cs:8,0)
--(axis cs:8,0)
--(axis cs:12,-0.0166666666666667)
--(axis cs:16,-0.009891098093474)
--(axis cs:24,0.025)
--(axis cs:32,0.0375)
--(axis cs:64,0.332755360847865)
--(axis cs:96,0.540392677388299)
--(axis cs:128,0.6890625)
--(axis cs:192,0.84122470282747)
--(axis cs:256,0.890133463782757)
--(axis cs:384,0.953663213777554)
--(axis cs:512,0.971950684189506)
--(axis cs:1024,0.990313791675303)
--(axis cs:2056,0.996221714898236)
--(axis cs:2056,0.998038985490869)
--(axis cs:2056,0.998038985490869)
--(axis cs:1024,0.995037770824697)
--(axis cs:512,0.981955565810494)
--(axis cs:384,0.969253452889112)
--(axis cs:256,0.934085286217243)
--(axis cs:192,0.885858630505863)
--(axis cs:128,0.7703125)
--(axis cs:96,0.620023989278367)
--(axis cs:64,0.460994639152135)
--(axis cs:32,0.175)
--(axis cs:24,0.133333333333333)
--(axis cs:16,0.047391098093474)
--(axis cs:12,0.05)
--(axis cs:8,0)
--cycle;

\path [draw=color0, fill=color0, opacity=0.2]
(axis cs:8,0)
--(axis cs:8,0)
--(axis cs:12,0)
--(axis cs:16,0)
--(axis cs:24,0)
--(axis cs:32,0)
--(axis cs:64,0)
--(axis cs:96,0)
--(axis cs:128,-0.0015625)
--(axis cs:192,0.0273582265342478)
--(axis cs:256,0.116756272817817)
--(axis cs:384,0.361822861022143)
--(axis cs:512,0.555704708687169)
--(axis cs:1024,0.866942699575634)
--(axis cs:2056,0.971327025649755)
--(axis cs:2056,0.976532896529233)
--(axis cs:2056,0.976532896529233)
--(axis cs:1024,0.877783862924366)
--(axis cs:512,0.590389041312831)
--(axis cs:384,0.41057297231119)
--(axis cs:256,0.135587477182183)
--(axis cs:192,0.0507667734657522)
--(axis cs:128,0.0046875)
--(axis cs:96,0)
--(axis cs:64,0)
--(axis cs:32,0)
--(axis cs:24,0)
--(axis cs:16,0)
--(axis cs:12,0)
--(axis cs:8,0)
--cycle;

\path [draw=blue, fill=blue, opacity=0.2, line width=0pt]
(axis cs:8,0)
--(axis cs:8,0)
--(axis cs:12,0)
--(axis cs:16,0)
--(axis cs:24,0)
--(axis cs:32,0)
--(axis cs:64,0)
--(axis cs:96,0)
--(axis cs:128,-0.00123638726168425)
--(axis cs:192,0.0261650757217499)
--(axis cs:256,0.113417637171073)
--(axis cs:384,0.352469565255268)
--(axis cs:512,0.55916493198984)
--(axis cs:1024,0.878259410351898)
--(axis cs:2056,0.968127062769003)
--(axis cs:2056,0.976522742678468)
--(axis cs:2056,0.976522742678468)
--(axis cs:1024,0.888342152148102)
--(axis cs:512,0.59278819301016)
--(axis cs:384,0.394926268078065)
--(axis cs:256,0.149082362828927)
--(axis cs:192,0.0436265909449168)
--(axis cs:128,0.00592388726168425)
--(axis cs:96,0)
--(axis cs:64,0)
--(axis cs:32,0)
--(axis cs:24,0)
--(axis cs:16,0)
--(axis cs:12,0)
--(axis cs:8,0)
--cycle;

\path [draw=green!50!black, fill=green!50!black, opacity=0.2, line width=0pt]
(axis cs:8,0)
--(axis cs:8,0)
--(axis cs:12,0)
--(axis cs:16,0)
--(axis cs:24,0)
--(axis cs:32,0)
--(axis cs:64,-0.003125)
--(axis cs:96,0.0193330140532182)
--(axis cs:128,0.0578125)
--(axis cs:192,0.269888283235263)
--(axis cs:256,0.497025800155369)
--(axis cs:384,0.707543518848677)
--(axis cs:512,0.828969418792883)
--(axis cs:1024,0.954053875343626)
--(axis cs:2056,0.987710974444924)
--(axis cs:2056,0.992736496372196)
--(axis cs:2056,0.992736496372196)
--(axis cs:1024,0.964500812156374)
--(axis cs:512,0.853061831207117)
--(axis cs:384,0.747143981151323)
--(axis cs:256,0.537349199844631)
--(axis cs:192,0.32594505009807)
--(axis cs:128,0.1078125)
--(axis cs:96,0.0410836526134485)
--(axis cs:64,0.00625)
--(axis cs:32,0)
--(axis cs:24,0)
--(axis cs:16,0)
--(axis cs:12,0)
--(axis cs:8,0)
--cycle;

\addplot [semithick, red]
table {%
8 0
12 0.0166666666666667
16 0.01875
24 0.0791666666666667
32 0.10625
64 0.396875
96 0.580208333333333
128 0.7296875
192 0.863541666666667
256 0.912109375
384 0.961458333333333
512 0.976953125
1024 0.99267578125
2056 0.997130350194552
};
\addlegendentry{HRR}
\addplot [semithick, color0, dashed]
table {%
8 0
12 0
16 0
24 0
32 0
64 0
96 0
128 0.0015625
192 0.0390625
256 0.126171875
384 0.386197916666667
512 0.573046875
1024 0.87236328125
2056 0.973929961089494
};
\addlegendentry{HRR w/ Proj}
\addplot [semithick, blue, dash pattern=on 1pt off 3pt on 3pt off 3pt]
table {%
8 0
12 0
16 0
24 0
32 0
64 0
96 0
128 0.00234375
192 0.0348958333333333
256 0.13125
384 0.373697916666667
512 0.5759765625
1024 0.88330078125
2056 0.972324902723736
};
\addlegendentry{VTB}
\addplot [semithick, green!50!black, dotted]
table {%
8 0
12 0
16 0
24 0
32 0
64 0.0015625
96 0.0302083333333333
128 0.0828125
192 0.297916666666667
256 0.5171875
384 0.72734375
512 0.841015625
1024 0.95927734375
2056 0.99022373540856
};
\addlegendentry{MAP-C}
\end{axis}

\end{tikzpicture}
        }
    }
     \subfigure[$d=8100$]{
         \centering
        \adjustbox{max width=0.3\textwidth}{
        \begin{tikzpicture}

\definecolor{color0}{rgb}{1,0.647058823529412,0}

\begin{axis}[
legend cell align={left},
legend style={fill opacity=0.8, draw opacity=1, text opacity=1, at={(0.03,0.97)}, anchor=north west, draw=white!80!black},
log basis x={2},
tick align=outside,
tick pos=left,
title={d=8100 Dimensions},
x grid style={white!69.0196078431373!black},
xlabel={Number of Vector Pairs and Distractors},
xmin=6.06168453445102, xmax=2713.43714878584,
xmode=log,
xtick style={color=black},
xtick={1,4,16,64,256,1024,4096,16384},
xticklabels={\(\displaystyle {2^{0}}\),\(\displaystyle {2^{2}}\),\(\displaystyle {2^{4}}\),\(\displaystyle {2^{6}}\),\(\displaystyle {2^{8}}\),\(\displaystyle {2^{10}}\),\(\displaystyle {2^{12}}\),\(\displaystyle {2^{14}}\)},
y grid style={white!69.0196078431373!black},
ylabel={P(Retrieval Error)},
ymin=-0.0761083856941961, ymax=1.04827609957812,
ytick style={color=black},
ytick={-0.2,0,0.2,0.4,0.6,0.8,1,1.2},
yticklabels={−0.2,0.0,0.2,0.4,0.6,0.8,1.0,1.2}
]
\path [draw=red, fill=red, opacity=0.2]
(axis cs:8,0.05)
--(axis cs:8,-0.025)
--(axis cs:12,0)
--(axis cs:16,-0.0125)
--(axis cs:24,-0.00600127712046063)
--(axis cs:32,0.0148532788563763)
--(axis cs:64,0.155120182932319)
--(axis cs:96,0.303514582151051)
--(axis cs:128,0.475729690120558)
--(axis cs:192,0.694773071403913)
--(axis cs:256,0.792942282147465)
--(axis cs:384,0.899362875837289)
--(axis cs:512,0.936050812396732)
--(axis cs:1024,0.984993718834727)
--(axis cs:2056,0.994661079890396)
--(axis cs:2056,0.997167713883923)
--(axis cs:2056,0.997167713883923)
--(axis cs:1024,0.989615656165273)
--(axis cs:512,0.956136687603268)
--(axis cs:384,0.929803790829378)
--(axis cs:256,0.839088967852535)
--(axis cs:192,0.753143595262754)
--(axis cs:128,0.558645309879442)
--(axis cs:96,0.417318751182282)
--(axis cs:64,0.216754817067681)
--(axis cs:32,0.0788967211436238)
--(axis cs:24,0.0810012771204606)
--(axis cs:16,0.025)
--(axis cs:12,0)
--(axis cs:8,0.05)
--cycle;

\path [draw=color0, fill=color0, opacity=0.2]
(axis cs:8,0)
--(axis cs:8,0)
--(axis cs:12,0)
--(axis cs:16,0)
--(axis cs:24,0)
--(axis cs:32,0)
--(axis cs:64,0)
--(axis cs:96,0)
--(axis cs:128,0)
--(axis cs:192,0)
--(axis cs:256,0.003125)
--(axis cs:384,0.0548626208871163)
--(axis cs:512,0.185443076093139)
--(axis cs:1024,0.644768473577772)
--(axis cs:2056,0.908820767540717)
--(axis cs:2056,0.918805691603252)
--(axis cs:2056,0.918805691603252)
--(axis cs:1024,0.674762776422228)
--(axis cs:512,0.214947548906861)
--(axis cs:384,0.0748248791128838)
--(axis cs:256,0.0140625)
--(axis cs:192,0)
--(axis cs:128,0)
--(axis cs:96,0)
--(axis cs:64,0)
--(axis cs:32,0)
--(axis cs:24,0)
--(axis cs:16,0)
--(axis cs:12,0)
--(axis cs:8,0)
--cycle;

\path [draw=blue, fill=blue, opacity=0.2, line width=0pt]
(axis cs:8,0)
--(axis cs:8,0)
--(axis cs:12,0)
--(axis cs:16,0)
--(axis cs:24,0)
--(axis cs:32,0)
--(axis cs:64,0)
--(axis cs:96,0)
--(axis cs:128,0)
--(axis cs:192,-0.00104166666666667)
--(axis cs:256,0.00078125)
--(axis cs:384,0.0554516321298571)
--(axis cs:512,0.177787463390605)
--(axis cs:1024,0.64472030376961)
--(axis cs:2056,0.915648952837301)
--(axis cs:2056,0.921705132765812)
--(axis cs:2056,0.921705132765812)
--(axis cs:1024,0.67520157123039)
--(axis cs:512,0.200728161609395)
--(axis cs:384,0.0752775345368095)
--(axis cs:256,0.009375)
--(axis cs:192,0.00208333333333333)
--(axis cs:128,0)
--(axis cs:96,0)
--(axis cs:64,0)
--(axis cs:32,0)
--(axis cs:24,0)
--(axis cs:16,0)
--(axis cs:12,0)
--(axis cs:8,0)
--cycle;

\path [draw=green!50!black, fill=green!50!black, opacity=0.2, line width=0pt]
(axis cs:8,0)
--(axis cs:8,0)
--(axis cs:12,0)
--(axis cs:16,0)
--(axis cs:24,0)
--(axis cs:32,0)
--(axis cs:64,0)
--(axis cs:96,0)
--(axis cs:128,0)
--(axis cs:192,0.0228819829391949)
--(axis cs:256,0.127980816845206)
--(axis cs:384,0.37388259075974)
--(axis cs:512,0.571808544667689)
--(axis cs:1024,0.872899329189622)
--(axis cs:2056,0.971280830660143)
--(axis cs:2056,0.978816445604448)
--(axis cs:2056,0.978816445604448)
--(axis cs:1024,0.897608483310378)
--(axis cs:512,0.606316455332311)
--(axis cs:384,0.42767990924026)
--(axis cs:256,0.170456683154794)
--(axis cs:192,0.0542013503941384)
--(axis cs:128,0)
--(axis cs:96,0)
--(axis cs:64,0)
--(axis cs:32,0)
--(axis cs:24,0)
--(axis cs:16,0)
--(axis cs:12,0)
--(axis cs:8,0)
--cycle;

\addplot [semithick, red]
table {%
8 0.0125
12 0
16 0.00625
24 0.0375
32 0.046875
64 0.1859375
96 0.360416666666667
128 0.5171875
192 0.723958333333333
256 0.816015625
384 0.914583333333333
512 0.94609375
1024 0.9873046875
2056 0.99591439688716
};
\addlegendentry{HRR}
\addplot [semithick, color0, dashed]
table {%
8 0
12 0
16 0
24 0
32 0
64 0
96 0
128 0
192 0
256 0.00859375
384 0.06484375
512 0.2001953125
1024 0.659765625
2056 0.913813229571985
};
\addlegendentry{HRR w/ Proj}
\addplot [semithick, blue, dash pattern=on 1pt off 3pt on 3pt off 3pt]
table {%
8 0
12 0
16 0
24 0
32 0
64 0
96 0
128 0
192 0.000520833333333333
256 0.005078125
384 0.0653645833333333
512 0.1892578125
1024 0.6599609375
2056 0.918677042801556
};
\addlegendentry{VTB}
\addplot [semithick, green!50!black, dotted]
table {%
8 0
12 0
16 0
24 0
32 0
64 0
96 0
128 0
192 0.0385416666666667
256 0.14921875
384 0.40078125
512 0.5890625
1024 0.88525390625
2056 0.975048638132296
};
\addlegendentry{MAP-C}
\end{axis}

\end{tikzpicture}
        }
    }

        \caption{Probability of a retrieval error (y-axis) when, given $n$ (x-axis) pairs of objects bound together and $n$ distractor items, the unbound concept vector is more similar to a distractor than the true original object. The values of $d$ are prefect squares due to a technical requirement of the VTB approach.  }
        \label{fig:p_error_retreival}
\end{figure}

For further edification about the capacity of each evaluated approach, we show in \autoref{fig:p_error_retreival} the probability of a retrieval error occurring as the number of $n$ items increases, with the standard deviation over 10 trials shown in the highlighted region. As can be seen, our improved HRRs and VTB have statistically indistinguishable performance, which was quite surprising, and may lead to further theory work around the limits of binding capacity in a fixed-length representation. 

In all cases we can see that while capacity at a threshold $t$ does increase linearly with dimension $d$ for the non-HRR approaches\footnote{It is possible naive HRRs would increase linearly given even larger values of $d$, but experimentation past that point is unreasonable.}. It is also worth noting that capacity is a fairly hard limit, with error increasing slowly until the capacity is reached, at which point the probability of error begins to increase rapidly with expanded set size. There are also other forms of VSA capacity that are beyond our current scope, especially when discussing mechanisms like RNNs built from VSA \cite{Frady2018}. Our results should not be taken as conclusive holistic descriptions of HRRs vs other VSAs, but are limited to the form of capacity we have discussed in this section and is most relevant to our application. 

In relation to our results in storing tens to hundreds of thousands of vectors, we note that our results in \autoref{sec:results} are based on learning to extract the correct objects, and the penalty term is based on a single averaged representation of all other concepts, which thus down weights any false-positive response due to noise of a single item. The capacity results we discuss in this section are with respect to \textit{any} of the original $\boldsymbol{z}_i$ distractors having a higher response, which requires $n$ brute force evaluations and is a harder scenario than what we required. 

\subsection{Capacity For Average Response Range}

The capacity question we have just walked through is for error free recognition of the true item as more similar than a set of $n$ distractors. However, our use of the HRR operations poses a mixed representation. In $J_p$ we perform extraction of the classes present, but $J_n$ relies on the average response value being accurate. $J_p$ requires on average less than 76 explicit items to be retrieved in all our datasets,  but $J_n$ is representing the average response over tens to hundreds of thousands of items. So while $J_n$ requires a ``larger'' capacity in some sense, it only requires the average response to be stable. 

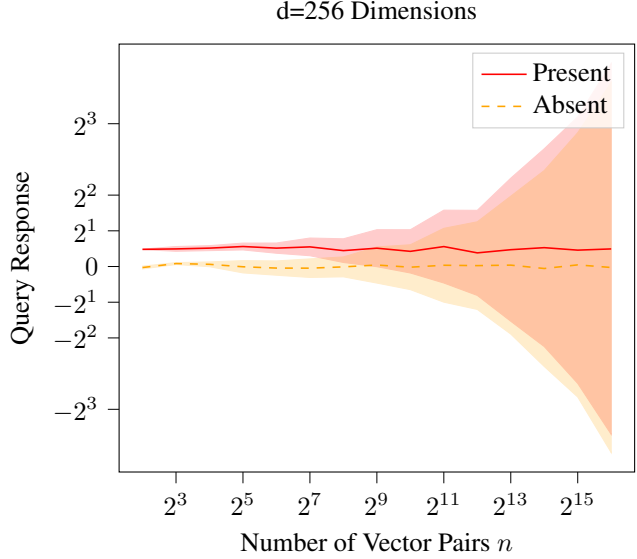
\begin{figure}[!h]
    \centering
    \adjustbox{max width=\textwidth}{
    \begin{tikzpicture}

\definecolor{color0}{rgb}{1,0.647058823529412,0}

\begin{axis}[
legend cell align={left},
legend style={fill opacity=0.8, draw opacity=1, text opacity=1, draw=white!80!black},
log basis x={2},
tick align=outside,
tick pos=left,
title={d=256 Dimensions},
x grid style={white!69.0196078431373!black},
xlabel={Number of Vector Pairs $n$},
xmin=2.46228882668983, xmax=106463.546095204,
xmode=log,
xtick style={color=black},
xtick={0.5,2,8,32,128,512,2048,8192,32768,131072,524288},
xticklabels={\(\displaystyle {2^{-1}}\),\(\displaystyle {2^{1}}\),\(\displaystyle {2^{3}}\),\(\displaystyle {2^{5}}\),\(\displaystyle {2^{7}}\),\(\displaystyle {2^{9}}\),\(\displaystyle {2^{11}}\),\(\displaystyle {2^{13}}\),\(\displaystyle {2^{15}}\),\(\displaystyle {2^{17}}\),\(\displaystyle {2^{19}}\)},
y grid style={white!69.0196078431373!black},
ylabel={Query Response},
ymin=-11.5121551561244, ymax=12.4622948968915,
ytick style={color=black},
ytick={-8,-4,-2,0,2,4,8},
yticklabels={\(\displaystyle {-2^{3}}\),\(\displaystyle {-2^{2}}\),\(\displaystyle {-2^{1}}\),\(\displaystyle {0}\),\(\displaystyle {2^{1}}\),\(\displaystyle {2^{2}}\),\(\displaystyle {2^{3}}\)}
]
\path [draw=red, fill=red, opacity=0.2]
(axis cs:4,0.975394279459809)
--(axis cs:4,0.948408503780664)
--(axis cs:8,0.851814540609625)
--(axis cs:16,0.890110745562508)
--(axis cs:32,0.928354647704242)
--(axis cs:64,0.743100391711344)
--(axis cs:128,0.603863898802381)
--(axis cs:256,0.226301348810196)
--(axis cs:512,-0.00981417108646654)
--(axis cs:1024,-0.362772236710976)
--(axis cs:2048,-0.922948795295039)
--(axis cs:4096,-1.61964181768852)
--(axis cs:8192,-3.06567425084248)
--(axis cs:16384,-4.48432272466327)
--(axis cs:32768,-6.54843365018626)
--(axis cs:65536,-9.40109116926281)
--(axis cs:65536,11.3725471672089)
--(axis cs:65536,11.3725471672089)
--(axis cs:32768,8.38263429687863)
--(axis cs:16384,6.59674155177493)
--(axis cs:8192,4.94969600098028)
--(axis cs:4096,3.14830536161776)
--(axis cs:2048,3.15712261091472)
--(axis cs:1024,2.06189205279631)
--(axis cs:512,2.06307866259277)
--(axis cs:256,1.55446134767606)
--(axis cs:128,1.59635277772908)
--(axis cs:64,1.31675699159442)
--(axis cs:32,1.31303102513785)
--(axis cs:16,1.16987804593325)
--(axis cs:8,1.11977547754987)
--(axis cs:4,0.975394279459809)
--cycle;

\path [draw=color0, fill=color0, opacity=0.2]
(axis cs:4,0.0203697460042666)
--(axis cs:4,-0.135643866665938)
--(axis cs:8,0.112003481414283)
--(axis cs:16,-0.0107543259708906)
--(axis cs:32,-0.358400592759214)
--(axis cs:64,-0.488001181916939)
--(axis cs:128,-0.613224867700499)
--(axis cs:256,-0.578096693637741)
--(axis cs:512,-0.927910598412111)
--(axis cs:1024,-1.29468836927191)
--(axis cs:2048,-1.99711681826739)
--(axis cs:4096,-2.4043636822716)
--(axis cs:8192,-3.78507343286106)
--(axis cs:16384,-5.59965915174215)
--(axis cs:32768,-7.3112291381803)
--(axis cs:65536,-10.4224074264418)
--(axis cs:65536,10.3130550716745)
--(axis cs:65536,10.3130550716745)
--(axis cs:32768,7.49468528502637)
--(axis cs:16384,5.37674052110369)
--(axis cs:8192,3.94014515904618)
--(axis cs:4096,2.50510651557063)
--(axis cs:2048,2.13812927115506)
--(axis cs:1024,1.21878902160031)
--(axis cs:512,1.09531760132112)
--(axis cs:256,0.53750306003244)
--(axis cs:128,0.427230979822506)
--(axis cs:64,0.312760687729026)
--(axis cs:32,0.329574439651925)
--(axis cs:16,0.256775802720938)
--(axis cs:8,0.21913195934201)
--(axis cs:4,0.0203697460042666)
--cycle;

\addplot [semithick, red]
table {%
4 0.961901391620236
8 0.985795009079748
16 1.02999439574788
32 1.12069283642105
64 1.02992869165288
128 1.10010833826573
256 0.890381348243127
512 1.02663224575315
1024 0.849559908042669
2048 1.11708690780984
4096 0.764331771964619
8192 0.9420108750689
16384 1.05620941355583
32768 0.917100323346185
65536 0.985727998973056
};
\addlegendentry{Present}
\addplot [semithick, color0, dashed]
table {%
4 -0.0576370603308356
8 0.165567720378147
16 0.123010738375023
32 -0.0144130765536447
64 -0.0876202470939565
128 -0.0929969439389967
256 -0.0202968168026503
512 0.0837035014545041
1024 -0.0379496738358001
2048 0.0705062264438366
4096 0.0503714166495177
8192 0.0775358630925628
16384 -0.111459315319226
32768 0.091728073423035
65536 -0.0546761773836478
};
\addlegendentry{Absent}
\end{axis}

\end{tikzpicture}
    }
    \caption{Distribution of the response to queries of the form $\boldsymbol{q}^\inv \sum_{i=1}^n \boldsymbol{a}_i \bind \boldsymbol{b}_i$ for cases where $\boldsymbol{q}$ is present or absent from the summation. The standard deviation of responses is shown in the shaded region, y-axis is symlog scale and x-axis is log scale. }
    \label{fig:response_dist}
\end{figure}

We can explore this in our data by looking at \autoref{fig:response_dist}, where we plot the mean and standard deviation of \textit{individual} responses. The solid lines correspond to the same results as presented in \autoref{fig:hrr_complex_proj_impact}, but we are looking at only the improved HRR, and showing the standard deviation of the individual responses that form the average.
 
Given this results there are multiple ways we could look at when the HRR response begins to ``fail''. If we look at when do the mean and standard deviation of the responses start to overlap for the present/absent cases, that starts around $n=512$ items. If we look at when the standard deviation starts to approach the other item's mean, that occurs around $n=1,536$. If we look at when the mean response begins to deviate away from the target value of 1/0, that does not start to occur until around $n=2^{16}= 65,536 $ (and is still very close, but larger values of $n$ are computationally expensive)! This stability of the average for large $n$ is an important component of our loss component $J_n = \cos{\left(\mathbf{m^{\dagger}} \bind \hat{\mathbf{s}}, \sum_{\boldsymbol{c}_p \in \mathcal{Y}^p} \boldsymbol{c}_p \right)}$ is implicitly working over an average response of all the negative labels. 

This shows that the distributional average around the desired response value for present/absent items is very stable, but the tails of the distribution do begin to grow as you try to pack more and more into the single representation. This validates further why we need to use a normalized response via the cosine similarity when extracting the present terms, but also how the $J_n$ term can function well despite the large symbolic query space. 

\section{Datasets} \label{sec:datasets}

All datasets and their source are given in \autoref{tab:exp_extreme_label_dataset_stats}.

\begin{table}[!h]
    \centering
  \caption{\textbf{Dataset Statistics} from \citet{Bhatia16}. The table describes the statistics of each dataset utilized for experiments and includes the number of features per sample, number of labels in each input sample, the diversity of the dataset represented through the average number of points per label and average number of labels in each sample.}
  \label{tab:exp_extreme_label_dataset_stats}
    \adjustbox{max width=\textwidth}{
    \begin{tabular}{c | c | c | c | c}
        \toprule
         \textbf{Dataset} & \textbf{Features} & \textbf{Labels} & \textbf{Avg. Samples per Label} & \textbf{Avg. Labels per Point}\\ \midrule
        \hline
        Mediamill \cite{snoek2006challenge} & 120 & 101 & 1902.15 & 4.38\\
        Bibtex \cite{katakis2008multilabel} & 1836 & 159 & 111.71 & 2.40 \\
        Delicious \cite{tsoumakas2008effective} & 500	& 983 & 311.61 & 19.03\\
        EURLex-4K \cite{leskovec2014snap} & 5000 & 3993 & 25.73 & 5.31\\
        \hline
        Wiki10-31K \cite{zubiaga2012enhancing} & 101938	& 30938 & 8.52 & 18.64\\
        Ama13K \cite{mcauley2013hidden} & 203882 & 13330 & 448.57 & 5.04\\
        Delicious-200K \cite{wetzker2008analyzing} & 782585	& 205443 & 2.29 & 75.54\\
        Amazon-670K \cite{leskovec2014snap} & 135909 & 670091 & 3.99 & 5.45 \\
        \bottomrule
    \end{tabular}
    }
\end{table}

\section{Additional Metrics} \label{eq:additional_metrics}

Next there is the DCG@k and PSDCG@k scores, which differ only by the inclusion of the $p_l$ term being absent / present respectively. PSDCG is shown below.  

\begin{equation*}
\text{PSDCG}@k := \sum_{l\in {\text{rank}}_k (\hat{\mathbf y})} \frac{\mathbf y_l}{p_l\log(l+1)}
\label{eq:exp_prop_dcg_k}
\end{equation*}

As recommended we use the normalized versions of each  giving us nDCG@k and PSnDCG@k, resulting in \cref{eq:nDCG} and \cref{eq:PSnDCG}.

\begin{equation}
\text{nDCG}@k := \frac{{\text{DCG}}@k}{\sum_{l=1}^{\min(k, \|\mathbf y\|_0)} \frac{1}{\log(l+1)}}
\label{eq:nDCG}
\end{equation}

\begin{equation}
\text{PSnDCG}@k := \frac{{\text{PSDCG}}@k}{\sum_{l=1}^{k} \frac{1}{\log(l+1)}}
\label{eq:PSnDCG}
\end{equation}

Across all experiments we see that results across different values of $k$ tend to be consistent. The pairings of  Precision@k and nDCG@k and PSprec@k and PSnDCG@k	 are highly correlated in all our results, and equivalent for $k=1$. For this reason we will show most results at $k=1$ for brevity, with larger tables of results in the appendix.

\section{Computer Resources} \label{sec:compute_resources}

Training was done primarily on a shared compute environment, but in general we had access to only one or two compute nodes at any given time. The main compute node used had a Tesla V100 with 32 GB of RAM, which could barely fit the Amazon-670K experiments during training. Going through all datasets to obtain results took approximately 2-weeks of compute time per model tested, and we have three models under evaluation. Combined with other experiments that did not pan out, we did not have the capacity to perform the 25+ runs that we would prefer to provide robust measures of variance in our results. We do report that spot checking smaller datasets like Bibtex that had large effect sizes consistently returned those large effect sizes. 

\section{Inference for XML with HRR}

We take a moment to be more explicit about how inference is done with HRRs to perform XML prediction, and also discuss further potential advantages that could be achieved given more software engineering effort. 

Given a network's prediction $\hat{\boldsymbol{s}} = f(\boldsymbol{x})$, inference can be done by simply iterating though all class HRR vectors $\boldsymbol{c}_1, \boldsymbol{c}_2, \ldots, \boldsymbol{c}_L$, and selecting  $\argmax_i \boldsymbol{c}_i^\top \boldsymbol{p}^\invA \bind \hat{\boldsymbol{s}}$ to determine that class $i$ is the top prediction of the network. To select the top-$k$ predictions, as is common in XML scenarios, one simply selects the top-$k$ largest dot products to be the predicted set. Or one can use a threshold of $\boldsymbol{c}_i^\top \boldsymbol{p}^\invA \bind \hat{\boldsymbol{s}} > 0.5$ to select the set of likely present classes. While this is \textit{not} a calibrated probability, this works out by the math of HRRs that a value being present should produce a dot product of $\approx 1$ and non-present values should produce a dot product of $\approx 0$. 

The above describes how inference is currently done in our code. We note that it could be further accelerated. This is because the inference formulation $\argmax_i \boldsymbol{c}_i^\top (\boldsymbol{p}^\invA \bind \hat{\boldsymbol{s}})$ is now a \textit{Maximum Inner-Product Search} (MIPS) problem, for which many algorithms have been designed to accelerate such queries \cite{pmlr-v55-luo16,Pratap2018,niculescu-mizil17a}. We have not incorporated these tools due to current freely available software not being well designed for our use case. This appears to be a purely software engineering problem, and beyond our current capacity to implement. For example, the MLPACK3 library\footnote{\url{https://www.mlpack.org/}} has MIPS algorithms that can perform the exact search for the top-$k$ items in expected $\bigO(\log n)$ time after building the index at cost $\bigO(n \log n)$. Our setup would allow such a construction, but the library is based on CPU only calculations. For the scale of datasets that are publicly available that we tested, the constant-factor speedup of a GPU is still faster than the $\bigO(
\log n)$ search. If we had access to private XML corpora with 100 million classes\cite{10.1145/3394486.3403368,10.1145/3289600.3290979}, we would expect this result to change. 
 
The only software we are aware of with GPU support for \textit{approximate} MIPS search is the FAISS library\footnote{\url{https://github.com/facebookresearch/faiss/wiki}}. While broadly useful, the library does not support the functionality we need to avoid significant overheads that make it slower than a brute force search in this case. First, the FAISS library requires keeping its own copy of all vectors $\boldsymbol{c}_1, \ldots \boldsymbol{c}_L$ in GPU memory. This is a non-trivial cost that can make it difficult for us to fit the model in memory at the same time, which is the case where such MIPS searches would prove advantageous. Our implementation does not require storing the symbols $\boldsymbol{c}_i$ in memory, because they can be re-constructed as needed based on a random seed. This makes the brute force search faster because it requires no additional memory accesses once $\boldsymbol{p}^\invA \bind \hat{\boldsymbol{s}}$ has been computed and stored in GPU memory. This makes our brute force considerably faster, and causes the FAISS implementation to have significant overhead for unneeded memory use in its normal index structure combined with explicitly storing  all  $\boldsymbol{c}_i$.

\section{HRR Model Runtime with XML-CNN}
In \cref{subsec_hrr_model_runtime}, we measured the performance of the baseline FFN (FC) and HRR-FFN (HRR-FC) and showed how its execution time decreases as the number of labels increase. The cost of a single forward pass through the network is lower than baseline because the size of the output layer is smaller. Similarly, we analyze the impact of output layer size reduction on the XML-CNN architecture \cite{liu2017deep}. We observe in \cref{tab:exp_xml_cnn_execution_time} that execution time reduces across larger datasets, but initially the optimization time is higher (amazoncat-12k). The optimization time accounts for both: (a) the time taken to compute the loss and (b) the time taken to calculate the gradients and update the network. 

\begin{table}[!h]
\centering
  \caption{\textbf{Model Execution \& Optimization Time}. Compare execution and optimization time for XML-CNN \cite{liu2017deep} and HRR-CNN. Execution time is the average time (seconds) to perform a forward pass and inference through the model for $1$ epoch of training. It corresponds to the throughput of the model. Similarly, optimization time includes the time to compute the loss and optimize the model. As observed, overall \textit{Train} time reduces as the number of labels in the dataset increases.}
  \label{tab:exp_xml_cnn_execution_time}
\begin{tabular}{@{}crcc@{}}
\toprule
\textbf{Dataset}                         & \multicolumn{1}{c}{\textbf{Model}} & \multicolumn{1}{c}{\textbf{Execution Time}} & \multicolumn{1}{c}{\textbf{Optimization Time}} \\ \midrule
\multirow{2}{*}{EURLex-4K}      & CNN               & \textbf{0.466}     &  \textbf{2.306}   \\
                                & HRR-CNN           & 0.467     &  3.657  \\ \cmidrule(l){2-4} 
\multirow{2}{*}{Wiki10-31K}     & CNN               & \textbf{0.630}     &  \textbf{2.665}   \\ 
                                & HRR-CNN           & 0.712     &   3.286  \\ \cmidrule(l){2-4}
\multirow{2}{*}{AmazonCat-12K}     & CNN            & 16.722     &  \textbf{83.305}  \\ 
                                & HRR-CNN           & \textbf{16.178}     &  117.098  \\ \cmidrule(l){2-4}                                
\multirow{2}{*}{Amazon-670K}    & CNN               & 239.48    &   734.694   \\
                                & HRR-CNN           & \textbf{122.665}   &  \textbf{301.376}  \\
\bottomrule
\end{tabular}
\end{table}

\begin{figure}[hbt!]
    \centering
    \includegraphics[width=\columnwidth]{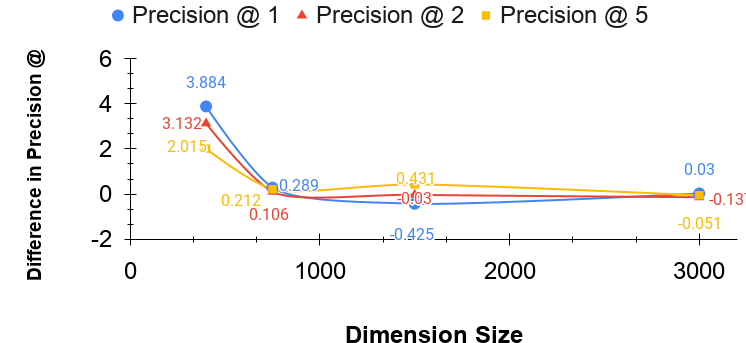}
\caption{Difference between the precision @ k for HRR-FC trained when $\boldsymbol{p}$ \& $\boldsymbol{m}$ vectors are updated. Positive values indicate a preference for fixed values, negative a preference for learned values. }
    \label{fig:exp_p_n_update_wiki10}
\end{figure}

\end{document}